\documentclass[11pt]{article}

\usepackage[preprint]{acl}

\usepackage{times}
\usepackage{latexsym}
\usepackage{multirow}
\usepackage[T1]{fontenc}
\usepackage[utf8]{inputenc}
\usepackage{microtype}
\usepackage{inconsolata}
\usepackage{graphicx}
\usepackage{xcolor}
\usepackage{amsmath}
\usepackage{amssymb}
\usepackage{enumitem}
\usepackage{booktabs}
\usepackage{array}
\usepackage{placeins}
\usepackage[capitalise,noabbrev]{cleveref}
\title{Same Context, Different Agents: A Behavioral Study of Semantic Layers,\\
Multi-Agent Orchestration and Memory Compaction in Data Agents}

\author{
  \normalfont
  Zhivar Sourati\textsuperscript{1,2} \quad
  Marianne Menglin Liu\textsuperscript{2} \quad
  Zheng Wang\textsuperscript{2} \\
  Rongguang Wang\textsuperscript{2} \quad
  Sai Ashish Somayajula\textsuperscript{2} \quad
  Tao Sheng\textsuperscript{2} \\
  Kyu J. Han\textsuperscript{2} \quad
  Sujith Ravi\textsuperscript{2} \quad
  Dan Roth\textsuperscript{2,3} \\[4pt]
  \textsuperscript{1}University of Southern California \quad
  \textsuperscript{2}Oracle AI \quad
  \textsuperscript{3}University of Pennsylvania
}

\begin{document}
\maketitle
\begin{abstract}
Answering questions over heterogeneous data means combining evidence from files, databases, documents and the web. Data agents automate this, but their success depends on context. Three context-engineering techniques (CETs) help by shaping it: semantic layers describe the data, orchestration splits work among agents, and compaction condenses interaction history. Existing studies validate each CET alone, leaving open how agent behavior shapes its effect. We present a comprehensive analysis of the three CETs, alone and combined, with seven large language models (LLMs) on 1,514 long-horizon tasks from five benchmarks, covering behavior, accuracy and cost. We find that data agents without CETs form three behavior groups: investigators, premature answerers that answer without inspecting data, and early stoppers that give up. Each CET delivers its known benefit, from more investigation to smaller prompts, but only the behavior group it matches turns that benefit into accuracy, such as premature answerers under orchestration. For investigators, the same CETs can hurt: the semantic layer stops them from looking, orchestration makes them overconfident, costing 5 percentage points, and compaction makes them repeat steps. Combining CETs yields smaller gains and losses than their sum. We derive guidance for choosing CETs by agent behavior, task demands and accuracy-cost trade-offs.
\end{abstract}

\section{Introduction}

\begin{figure}[t]
  \centering
  \includegraphics[width=\columnwidth]{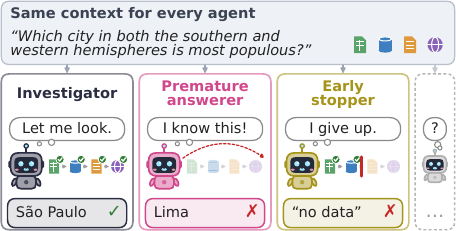}
\caption{\textbf{Same context, different agents (illustration).} Given the same question and data, an \textbf{investigator} checks every source and answers from the data, a \textbf{premature answerer} answers without looking, and an \textbf{early stopper} gives up partway. }
  \label{fig:teaser}
\end{figure}

The goal of data analysis is to turn raw data into answers. Analysts start from a question but must find the relevant sources among many files, tables, documents and web pages \citep{lai2026kramabench,zhang2026dataclawbench,wang2026fdabench,wu2026benchmarking}, read data that is heterogeneous across and within sources, with irregular layouts, entity names in multiple languages, obfuscated columns and metrics defined elsewhere \citep{ma2026dab,li2026dataspace,lei2025spider}, align it and compute a result. Agents built on large language models (LLMs) that reason and act with tools \citep{yao2022react,jimenez2023swe,liu-etal-2026-toolscope} are increasingly applied to this problem \citep{hong2025data,jing2025dsbench} and complete ever longer tasks \citep{kwa2025measuring}, yet they remain unreliable \citep{jing2025dsbench,lai2026kramabench}. Much of the difficulty comes from the model's context. The data is too large to show in full, so the agent must decide what to look at; its history grows with every step; and models use long inputs unevenly, overlooking information in the middle of a long prompt and degrading as it grows even when it holds what they need \citep{liu2024lost,du2025context}. The result is an agent that forgets what it found, states values it never computed or acts without knowing where to look. Given the same context, agents fail differently: one checks every source, another answers without looking, and a third gives up partway (\Cref{fig:teaser}).

\begin{figure*}[t]
  \centering
  \includegraphics[width=\textwidth]{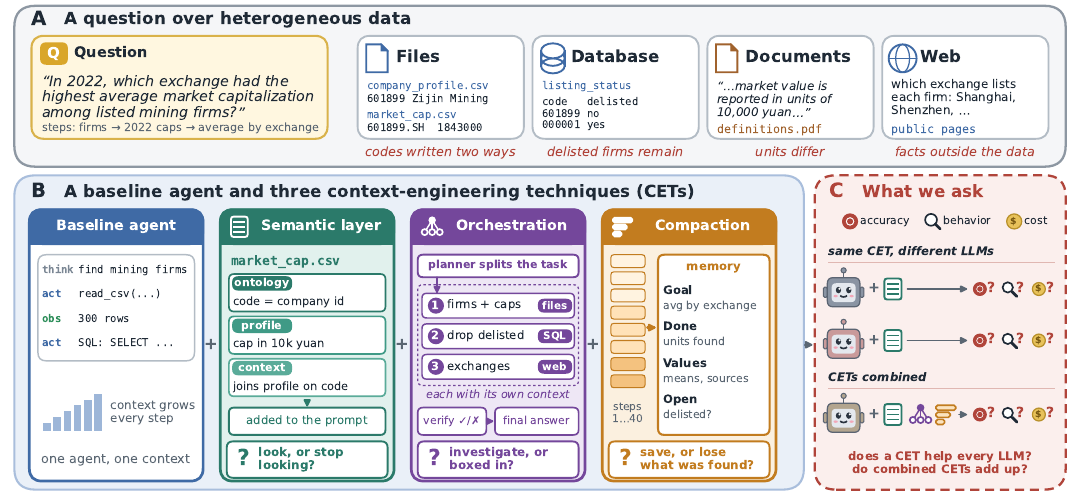}
  \caption{\textbf{Data agents and the context-engineering techniques that help them.} Data agents answer questions over heterogeneous data, and context-engineering techniques (CETs) help by shaping the context the model works from: a semantic layer describes the data, orchestration splits the work and compaction condenses the history. What we do not yet know is how each changes what different LLMs do, alone and combined.}
  \label{fig:overview}
\end{figure*}

Context engineering designs what an LLM conditions on at each step \citep{mei2025survey}. Three context-engineering techniques (CETs) target these problems (\Cref{fig:overview}B). A \emph{semantic layer} supplies a curated description of the data \citep{sequeda2024benchmark,rumiantsau2026semantic, chong2026evoontology}. \emph{Orchestration} distributes work among a planner and specialist agents, each with its own context \citep{wu2023autogen}. \emph{Compaction} replaces the growing history with a bounded working memory \citep{packer2023memgpt,zhou2026mem1,kang2025acon}. Each is usually studied alone, in a setting built for it, and judged by its average accuracy effect, although all three edit the same finite prompt (\Cref{sec:related}). Such studies tell whether a CET helps, but not what it changes in how an agent works, for which models or when CETs meet.

Practitioners need to know which failure their agent has, which CET repairs it and when it breaks. Asked the question in \Cref{fig:teaser}, GPT-5.4 opens the file that answers it and replies S\~ao Paulo; given a semantic layer naming that file and saying to verify values with tools, it repeatedly replies ``Lima'' without any tool, as if the layer gave it the answer. GPT-5.5 answers which of four exchanges had the highest average market value among mining firms in 5-7 calls; with compaction it records the same answer by call nine, then ``verifies'' it for about 300 more actions. Alone, GPT-6 Astra correctly answers in four calls whether any metro area had more identity-theft than fraud reports; with orchestration, its verifier refuses a table labeled ``Fraud and Other'' as a stand-in for fraud and after 41 calls reports the question cannot be determined.

Such cases raise four research questions (RQs; \Cref{fig:overview}B-C). \textbf{RQ1}: How do LLMs behave and fail as data agents without CETs? \textbf{RQ2}: Does each CET help every model, and at what cost? \textbf{RQ3}: What does each CET change in how an agent works: does a semantic layer make it look at the data (inspect it with tools) or stop looking, does orchestration make it investigate or box a strong one in, and does compaction save context and money or lose what was found? \textbf{RQ4}: Do CETs add up?

We study these questions where CETs are pushed to their limits: long-horizon tasks whose evidence is spread over files, databases, documents and the web, where the agent must choose what to inspect. Each CET is switched on or off on top of a \emph{baseline agent}, a single tool-using agent with a growing context, and we test all eight combinations with seven LLMs on five benchmarks (1,514 tasks), with model, tools, limits and prompt fixed. A CET's effect is not a property of the CET but of its match with the agent's behavior: the same CET can rescue one model and degrade another, and what it was designed to improve, such as tool use, calls and prompt size, can improve while answers get worse. Orchestration gives models that answer before looking 23 points at four times the cost; the layer gives models that give up a larger gain than orchestration at a fraction of its cost; compaction cuts their give-ups on the longest tasks; and no CET raises the accuracy of agents that already investigate. Choosing a CET is thus a question about the agent, not a default.

Our contributions are: (1) a controlled evaluation of data agents under every combination of semantic layers, orchestration and compaction, which prior work evaluates one at a time, measuring accuracy, behavior, cost and each benchmark's process measures; (2) a behavioral analysis of every run's trace that sorts the seven LLMs into three behavior groups and shows that each CET helps the groups whose habit it repairs while harming the investigators, and that combined CETs yield smaller gains and losses than their sum; (3) an analysis of wrong answers showing that selecting the wrong slice of data remains the most common cause after the agent inspects the data, under every combination of CETs; and (4) practical guidance for choosing and combining CETs by agent behavior, task demands and the accuracy-cost trade-off.

\section{Framework}
\label{sec:framework}

An agent answers a question by acting on files, databases, documents and, where allowed, the web. Switching each CET on or off on top of a baseline agent gives eight configurations that share the same model, tools, decoding settings and budget of steps and calls (\Cref{fig:overview}B; \Cref{app:config}).

\paragraph{Baseline agent.} A single tool-using agent \citep{yao2022react}. At each step it writes a thought and issues one action (Python, SQL, a file operation, or a web or document search), observes the result and decides what to do next, until it answers. It therefore already plans, inspects any source, tests intermediate results with code and corrects its course when an action fails or a result looks wrong. Its prompt tells it that computed values must come from executed code over the data, not from the question's wording, and to keep working until it has computed the answer. The CETs do not add functions the baseline agent lacks; they change its context and how its work is organized.

\paragraph{Semantic layer.} A description of the data, written offline once per benchmark by an LLM from the data, with no answers or task-specific hints. For every source it records the \emph{ontology} (what the source, its tables and columns are), the \emph{data profile} (value samples, units, missing values, parsing recipes) and the \emph{contextualization} (join keys, related sources, ambiguous names). The entries most relevant to the question go into the prompt before the first step, and the agent must verify them.

\paragraph{Orchestration.} A planner without tools splits the question into subtasks and delegates them to specialist agents for files, Python, SQL, or web and document search. Each uses the baseline agent's tools in its own context and returns a condensed result. A verifier accepts each result or sends the work back. With the layer on, the planner passes the relevant entries to each specialist.

\paragraph{Compaction.} The agent carries a bounded working memory instead of the full history, rewritten after every step: the goal, what is done, computed values with their evidence and open items.

\section{Experimental Setup}
\label{sec:setup}

\paragraph{Benchmarks.} We use five benchmarks with 1,514 tasks in total, counting FDABench's report and choice tasks as two. \textbf{KramaBench} \citep{lai2026kramabench} (104 tasks) asks questions over data lakes of many files with irregular layouts. \textbf{DataClaw} \citep{zhang2026dataclawbench} (492 tasks) poses consulting questions over noisy enterprise tables and policy texts with entity names in two languages, scored by an LLM judge. \textbf{FDABench} \citep{wang2026fdabench} joins a database with documents and web evidence, as 218 report tasks scored by a rubric and 320 multiple-choice tasks. \textbf{HybridDeepResearch} (HDR; \citealt{wu2026benchmarking}; 380 tasks) gives the agent no schema (it must discover tables by querying) and chains clues across the database and the web.

\paragraph{Models.} GPT-5.4, GPT-5.5 and GPT-6 Astra \citep{openai2026gpt54,openai2026gpt55,openai2026astra}, Claude Sonnet 5.5 and Opus 5.5 \citep{anthropic2026opus55,anthropic2026sonnet55}, Gemini 2.5 Flash \citep{comanici2025gemini} and the open-weight GLM-4.7-Flash \citep{zeng2025glm}, served locally, at provider defaults (\Cref{app:effort}); GPT-5.5 wrote the layers.

\paragraph{Metrics and analysis.}
Accuracy is each benchmark's own score (\Cref{app:published}). Cost is dollars per task at list prices (\Cref{app:cost}). A CET's effect is the change when it is switched on, averaged over the settings of the other two. Accuracy effects are in percentage points with 95\% bootstrap intervals over tasks, and group effects average their models.

\paragraph{Behavior.} From each run's trace (\Cref{app:behavior,app:annotation}) we measure: \textbf{Acts}, runs at least one tool; \textbf{Answer grounded} (\textbf{Grounded}), the submitted value appears in a tool output; \textbf{Covers sources} (\textbf{Coverage}), share of needed source types touched; \textbf{Persists}, three or more actions; \textbf{Recovers}, share of failed calls followed by a success; \textbf{No repeats}, no identical action twice; \textbf{Flags uncertainty} (\textbf{Hedges}), a hedged answer (annotated by an LLM); \textbf{No give-up}, does not end saying no relevant data exists; \textbf{Answers}, ends with an answer; and \textbf{Long loops}, 100 or more model calls.
\section{Results}
\label{sec:results}

We answer the research questions in turn: how the models behave as baseline agents (RQ1, \Cref{sec:plain}), whether each CET helps every model (RQ2, \Cref{sec:overview}), what each changes in how an agent works (RQ3, \Cref{sec:orch,sec:layer,sec:compress}) and whether the three add up (RQ4, \Cref{sec:combine}).

\subsection{Three Groups of Baseline Agents}
\label{sec:plain}

\begin{figure}[t]
  \centering
  \includegraphics[width=\columnwidth]{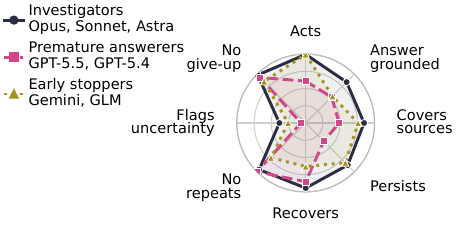}
  \caption{\textbf{How the baseline agent behaves, by behavior group.} Each axis is the share of a group's baseline runs that show a behavior, from 0 at the center to 100\% at the outer ring (mean of the group's models). Investigators look and persist, premature answerers often answer without acting, and early stoppers act but give up and repeat themselves.}
  \label{fig:profile}
\end{figure}

Under identical tools and instructions, the seven baseline agents fall into three behavior groups, \emph{investigators}, \emph{premature answerers} and \emph{early stoppers}, which clustering recovers exactly (\Cref{fig:profile}). Investigators lead every benchmark (\Cref{tab:plain}). For each group, our analyses rule out the obvious explanation.

\noindent\textbf{Investigators' accuracy advantage is linked to grounding.} Opus, Sonnet and Astra act in almost every run, ground most answers in tool outputs and rarely give up. Their edge is not knowledge: overall, answers without any tool are right in 7\% of runs, against 44\% after looking (\Cref{app:behavior}).

\noindent\textbf{Premature answering can persist despite reasoning.} GPT-5.5 and GPT-5.4 often reply without any tool (on 54\% and 24\% of tasks), although the prompt tells them to compute from the data. When GPT-5.5 does act, its accuracy is more than twice as high, so this is a habit, not a lack of ability or of thought: more reasoning does not reliably remove it, while orchestration, which demands evidence, does (\Cref{app:effort,sec:orch}).

\noindent\textbf{Early stoppers abandon solvable tasks despite substantial effort.} Gemini and GLM act, but give up with ``no relevant data'' in about 15\% of runs and repeat an identical action in 28-29\%. They do not give up for lack of data or effort: when they do, they have usually touched every source type the task needs, after three times as many actions as an investigator's typical run, and at least one investigator solves 80\% of those tasks (\Cref{app:fail}).

After looking, all groups fail alike, mostly by selecting the wrong data slice or misreading documents and web evidence, rarely by arithmetic, and no CET removes these causes (\Cref{fig:causes}).

\begin{table}[t]
\centering
\footnotesize
\setlength{\tabcolsep}{3pt}
\resizebox{\columnwidth}{!}{%
\begin{tabular}{lccccc}
\toprule
 & Krama & DataC. & FDA rep. & FDA ch. & HDR \\
\midrule
Opus 5.5 & 82 & 71 & 60 & 70 & 61 \\
GPT-6 Astra & 81 & 72 & 59 & 67 & 69 \\
Sonnet 5.5 & 79 & 66 & 46 & 56 & 61 \\
\midrule
GPT-5.5 & 22 & 63 & 10 & 3 & 9 \\
GPT-5.4 & 37 & 28 & 28 & 4 & 8 \\
\midrule
Gemini 2.5 Flash & 50 & 22 & 24 & 8 & 19 \\
GLM-4.7-Flash & 21 & 15 & 29 & 3 & 6 \\
\bottomrule
\end{tabular}}
\caption{\textbf{Baseline agent accuracy} in \% on each benchmark's own metric, with rows by behavior group (investigators, premature answerers, early stoppers).}
\label{tab:plain}
\end{table}

\subsection{CET Benefits Depend on Agent Behavior and Task Demands}
\label{sec:overview}

\begin{figure*}[t]
  \centering
  \includegraphics[width=\textwidth]{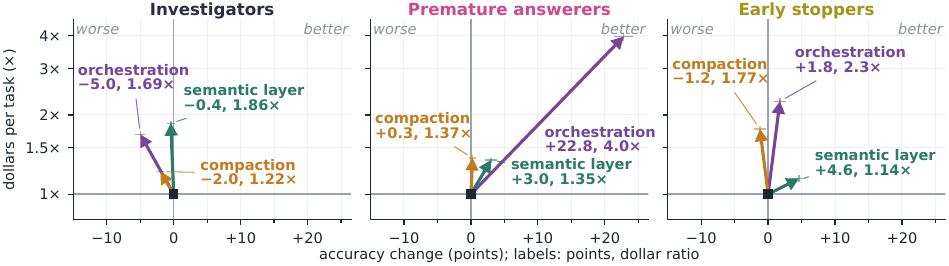}
  \caption{\textbf{Accuracy-cost trade-offs depend on the behavior group.} Each arrow goes from the baseline agent (square) to the agent with one CET: accuracy change in points and dollars per task as a multiple of the baseline's. Orchestration gives the largest gain for premature answerers and the semantic layer for early stoppers; for investigators every CET lowers accuracy or leaves it unchanged, and every CET raises the cost.}
  \label{fig:arrows}
\end{figure*}

\noindent\textbf{Every CET costs more, and what it returns depends on the group.}
Orchestration and the semantic layer help the weaker groups (\Cref{fig:arrows}). For premature answerers, orchestration gives the largest gain we measure, 23 points, at four times the dollars. For early stoppers, the layer is cheaper and gives them over twice orchestration's gain (4.6 against 1.8 points), at 1.1 rather than 2.3 times the dollars. Compaction does its job, keeping every prompt small, and pays off where history grows longest; averaged over tasks it changes accuracy by at most two points. The strongest models are where the expectation fails: for investigators, every CET lowers accuracy or leaves it unchanged, at a higher cost. CETs built to help an agent with its context make the strongest models worse.

\begin{figure*}[t]
  \centering
  \includegraphics[width=\textwidth]{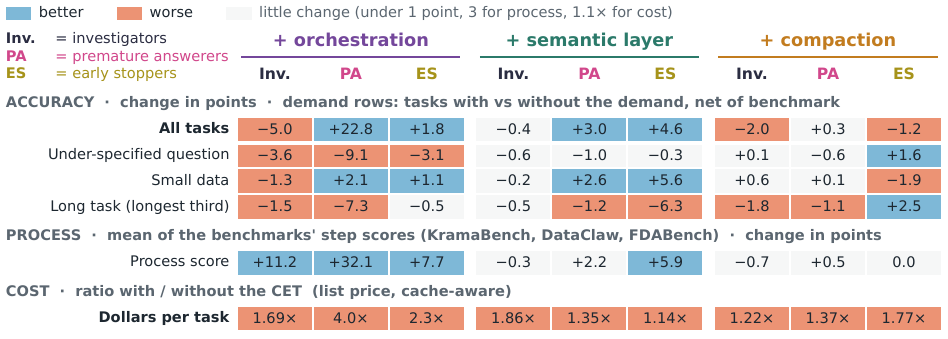}
  \caption{\textbf{Task demands modify CET effects, and process gains can accompany accuracy losses.} Rows show each CET's effect on accuracy (overall and within-benchmark contrasts between tasks with and without each demand), process score (the mean of three benchmarks' step scores), and dollars per task. Blue indicates improvement and red degradation; white marks effects below the display thresholds of one accuracy point, three process-score points, and 1.1 times the cost. \Cref{fig:scorecard-full} reports every demand, each benchmark's process measure, and which changes are significant.}
  \label{fig:scorecard}
\end{figure*}

\noindent\textbf{Each CET also responds to a different side of the task.}
LLMs label what each task demands, such as an under-specified question, small data or a long task, and we compare tasks with and without each demand within a benchmark (\Cref{fig:scorecard}; \Cref{app:demand-profile}). Orchestration responds to the question: it loses accuracy when the question is under-specified, as orchestrated agents stop stating the assumptions it needs. The layer responds to the data: it helps most on small, lightly messy data, and less when three or four sources must be combined. Compaction responds to the length of the work: its effect depends on task length (\Cref{sec:compress}), and the longest tasks also shrink the other CETs' gains.

\begin{figure*}[t]
  \centering
  \includegraphics[width=\textwidth]{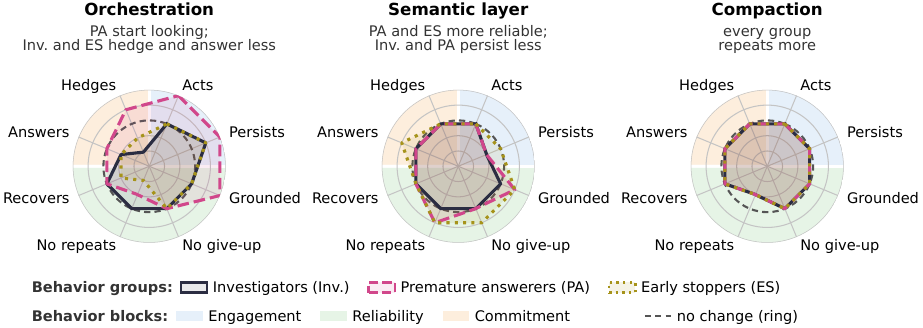}
  \caption{\textbf{How each CET changes what agents do.} For each group, a line outside the dashed ring means the CET makes a behavior more common and a line inside it less common; behaviors are grouped into engagement, reliability and commitment. Orchestration raises engagement, the semantic layer raises reliability at the cost of engagement, and compaction lowers reliability; \Cref{fig:fingerprint} gives every measure.}
  \label{fig:shifts}
\end{figure*}

\noindent\textbf{Process scores can rise while accuracy falls.}
The benchmarks' process scores, the share of expected steps a run takes, rise most with orchestration (\Cref{fig:scorecard}), but do not tell whether answers improve: orchestration raises investigators' process score by 11 points while their accuracy falls by 5. Neither shows how an agent works, so we measure behavior (\Cref{fig:shifts,fig:fingerprint}), in three blocks: engagement (Acts, Coverage, Persists), reliability (Grounded, No give-up, No repeats, Long loops, Recovers) and commitment (Answers, Flags uncertainty).

\noindent\textbf{Behavior shows what the scores hide.}
Orchestration raises engagement, most for premature answerers, who act, cover sources and persist in 36-50 points more runs, and its cost falls on commitment, which neither score sees: investigators, whose process score rises while their accuracy falls, flag uncertainty and end with an answer less often, as do the early stoppers. The layer raises reliability at the expense of engagement: runs repeat and loop less in every group but persist less, so investigators' unchanged scores hide that they look less. Compaction barely changes either score but adds repeated actions in all groups. A CET helps when it changes the behavior the agent lacks; otherwise it changes the agent without improving it.

\subsection{Orchestration Helps Premature Answerers but Can Harm Investigators}
\label{sec:orch}

\noindent\textbf{Orchestration delivers what it promises.}
The agent uses more tools, reaches more sources and persists longer, and for premature answerers this pays off: GPT-5.5 gains 37-47 points on benchmarks where it often answers without a tool. The gain is largest where the baseline agent is weakest, as \citet{kim2025towards} also find, but it follows the habit more than the weakness: GPT-5.5 and Gemini start from similar accuracy (21\% and 25\%), yet orchestration gives GPT-5.5 35 points and Gemini less than one, because GPT-5.5 does not look and Gemini does (\Cref{app:demand-profile}).

\noindent\textbf{The gain is not a larger budget.}
A single GPT-5.5 agent forced to act before answering, or to spend as many calls as orchestration, closes at most 37\% of the gap (\Cref{fig:compute}): it runs one action, then repeats the answer it gave before looking. What orchestration adds is structure: a planner that asks for evidence, and specialists that must return it.

\noindent\textbf{Orchestration also has a cost.}
For investigators, it lowers accuracy by 5 points at 1.7 times the dollars, mostly on tasks that chain a database with documents and web search (6-12 points), and not for lack of data: it usually reaches every needed source (\Cref{app:orch}).

\noindent\textbf{It changes how an answer is judged and stated.}
The verifier approves 87\% of the investigators' wrong answers, or, for Astra, rejects usable evidence until the planner abstains on 15.5\% of HDR tasks, and investigators hedge far less. Orchestration makes a careful agent more confident and less thorough.

\begin{figure*}[!tb]
  \centering
  \includegraphics[width=\textwidth]{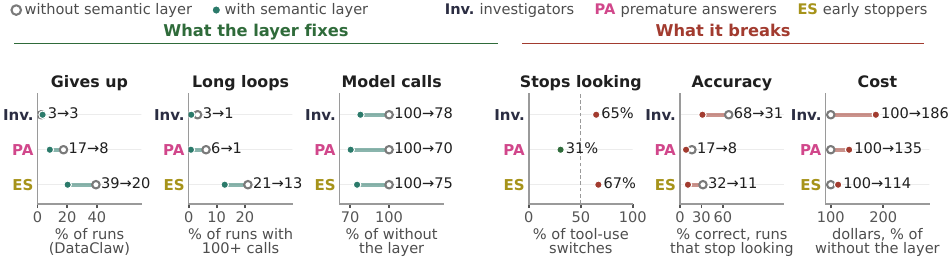}
  \caption{\textbf{The semantic layer cuts give-ups, long loops and calls, but raises the bill and makes investigators stop looking.} It helps most the agents that give up, halving their give-ups and loops at little extra cost. Investigators instead stop using tools on most runs whose tool use changes, and their accuracy there falls from 68\% to 31\%.}
  \label{fig:layersb}
\end{figure*}

\subsection{Semantic Layers Help Early Stoppers but Can Discourage Investigation}
\label{sec:layer}

\noindent\textbf{For the weaker groups, the layer works.}
The semantic layer is built to help in three ways: the ontology says what each source is, the data profile how to read it, and the contextualization how sources relate. It names candidate sources, contains no answers and tells the agent to check every entry. It works as built (\Cref{fig:layersb}): told that an answer exists and where it lives, the models that give up most halve their give-ups on DataClaw, and early stoppers ground more answers and loop less. Each part does what it is built for (\Cref{app:layer}): with the ontology, agents open the source it lists first in 27-42\% of runs, against 4-13\%; with the data profile, 16-20 points fewer runs fail to read a file; and the contextualization's join keys are used where joins are needed.

\noindent\textbf{Fewer calls do not mean lower cost.}
A layer is also expected to make an agent cheaper, and every group needs 22-30\% fewer calls; but the layer is in every prompt, so dollars rise 1.1-1.9 times, and only for the early stoppers is the gain nearly free.

\noindent\textbf{The layer also has a side effect: told what the data contains, the investigators look less.}
Opus stops using tools in 73 runs where its baseline agent used them and starts in only 11, and on the investigators' runs that stop looking, accuracy falls from 68\% to 31\%. The layer is rarely wrong: of 100 investigator answers that went from right to wrong with the layer, it misled the agent in 2, while in 25 the agent stopped checking. Nor does the writer matter much: a layer written by a 3B model gives nearly the same gain as one by GPT-5.5 (\Cref{fig:author}). The layer is a guide, and the investigators treat it as the answer. If a layer correctly says a sales table lists a price per product, for instance, the agent reads that column and answers, where without the layer it would open the table and notice a second column of discounted prices. Although told to verify every entry, they stop checking, the checking that sets them apart (\Cref{app:layer}). For an agent that looks too little the layer is a map; for one that already looks, it replaces looking \citep[cf.][]{chong2026evoontology}.

\begin{figure*}[!tb]
  \centering
  \includegraphics[width=\textwidth]{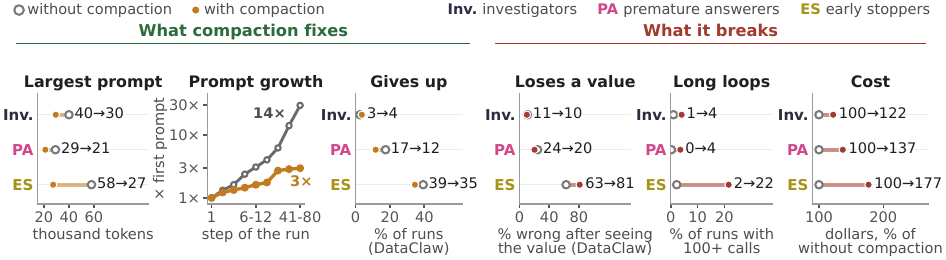}
  \caption{\textbf{Compaction limits prompt growth but increases cost and repetition.} ``Loses a value'' measures incorrect final answers after the correct value appeared in a tool output on DataClaw.}
  \label{fig:compsb}
\end{figure*}

\subsection{Compaction Bounds Context but Can Increase Cost and Repetition}
\label{sec:compress}

\noindent\textbf{Compaction delivers the benefit it is known for.}
Each prompt stays near three times the size of the first prompt (\Cref{fig:compsb}), whereas without compaction the full transcript grows to fourteen times. GLM-4.7-Flash, with its short context window, no longer has to truncate its history when it orchestrates. The working memory keeps what matters: after every DataClaw rewrite, most models kept 92-100\% of computed values.

\noindent\textbf{Its accuracy effect depends on task length.}
Where the history grows longest, compaction does most for the agents that lose track of it: on tasks that take other models longest, the early stoppers give up less and do 2.5 points better than on their other tasks, and GPT-5.4 gives up less. On short tasks, with little to compress, early stoppers lose 8 points, as each rewrite risks losing what was found.

\noindent\textbf{It also has costs: a smaller prompt is not a smaller bill.}
Every step adds a call that rewrites the memory, so total input barely changes (4-17\% more tokens) while output, the costliest token, grows 2.3 to 3.5 times, and dollars rise 1.2 to 1.8 times.

\noindent\textbf{And some agents loop.}
A model that distrusts its result keeps checking it: 81\% of premature answerers' repeats re-check a value still in memory, and 42\% of GPT-5.5's DataClaw runs reach 100 or more calls (none without it) at eight times the cost.

\subsection{Combining CETs Offers Limited Added Gains but Can Mitigate Harms}
\label{sec:combine}

\noindent\textbf{Combined effects are smaller than the sum of their parts.}
One might expect CETs to help each other, each fixing what others miss. For accuracy they mostly do not (\Cref{tab:cells}): they fix the same behaviors, so each helps less when another is on, and their harms do not add up either (\Cref{app:combine}). Each pair shows a different side of this.

\noindent\textbf{The layer adds little to orchestration.}
The layer and orchestration both help agents find data. One might expect the layer to make orchestration cheaper, since its specialists search less, and they make fewer calls; but every prompt carries the layer, so the cost rises while accuracy barely does.

\noindent\textbf{Orchestration and compaction combine worst.}
Under orchestration, compaction makes more runs loop (10 more points of runs reach 100 or more calls, against 6 without it) and cuts orchestration's gain in grounded answers from +12 to +4 points.

\noindent\textbf{The layer and compaction work well together.}
Compaction rewrites the agent's memory after every step, which makes runs loop and lose grounding; the layer is not part of that memory, and with it compaction's rise in long runs falls from 9 to 6 points, its multiplier on calls from 4.1 to 3.0, and under orchestration its drop in grounded answers from 6 points to under 1, most for models that loop most. The layer thus combines best: it adds little where another CET already fixes the same habit, but makes compaction cheaper and safer.

\noindent\textbf{In practice, a second CET should repair what the first leaves.}
Start with the CET matching the agent's habit, and add another only for a habit it leaves or a behavior it causes: leave an agent that looks alone; give one that does not orchestration, plus a layer if it also gives up; use compaction for long runs, with the layer. Measure each combination: its accuracy and cost cannot be read from its parts.

\begin{table}[t]
\centering
\footnotesize
\setlength{\tabcolsep}{2pt}
\resizebox{\columnwidth}{!}{%
\begin{tabular}{lcccccccc}
\toprule
 & baseline & +C & +S & +S+C & +M & +M+C & +M+S & +M+S+C \\
\midrule
Opus 5.5 & \textbf{68.8} & 65.2 & 65.6 & 65.3 & 61.9 & 64.0 & 61.8 & 63.1 \\
GPT-6 Astra & \textbf{69.6} & 65.1 & 69.6 & 65.7 & 61.0 & 56.4 & 61.0 & 56.4 \\
Sonnet 5.5 & \textbf{61.3} & 57.7 & 61.2 & 61.8 & 60.0 & 58.4 & 57.0 & 56.0 \\
\midrule
GPT-5.5 & 21.4 & 20.8 & 23.7 & 23.9 & \textbf{58.1} & 55.6 & 57.8 & 56.4 \\
GPT-5.4 & 21.0 & 27.0 & 29.2 & 31.3 & 37.2 & 36.4 & \textbf{40.1} & 39.3 \\
\midrule
Gemini 2.5 Flash & 24.6 & 22.3 & 27.0 & 26.0 & 24.7 & 20.9 & \textbf{29.2} & 27.7 \\
GLM-4.7-Flash & 14.9 & 14.8 & 21.8 & 19.9 & 19.0 & 18.9 & 22.1 & \textbf{23.0} \\
\bottomrule
\end{tabular}}
\caption{\textbf{Accuracy of every combination of CETs} (\%, mean of the five benchmarks; M: orchestration, S: semantic layer, C: compaction). The best combination depends on the group: no CET for investigators, orchestration, mostly with the layer, for the others.}
\label{tab:cells}
\end{table}

\section{Related Work}
\label{sec:related}

\paragraph{Data agent benchmarks.} LLM evaluation on data has moved from one query over one database \citep{lei2025spider,chen2024beaver} to agents that find, read and combine evidence across data lakes of irregular files \citep{lai2026kramabench}, enterprise tables and policy texts \citep{zhang2026dataclawbench}, databases joined with documents and the web \citep{wang2026fdabench,wu2026benchmarking}, and heterogeneous workspaces \citep{ma2026dab,li2026dataspace}. Even strong agents are far from reliable \citep{jing2025dsbench}: the best published KramaBench agent scores 56\%, humans 77\% \citep{lai2026kramabench}. They also trace part of the difficulty to how agents behave, not what they can reach: giving an agent only the gold files raises KramaBench accuracy by 0-7\% \citep{lai2026kramabench}, DataClaw finds models stopping early on tasks they fail \citep{zhang2026dataclawbench}, and reasoning models can cost more without answering better \citep{wang2026fdabench}.

\paragraph{Context-engineering techniques.} A line of work adds techniques that manage what the model sees. A semantic layer describes the data, and such descriptions raise the accuracy of question answering over databases \citep{sequeda2024benchmark,rumiantsau2026semantic,somayajula2026soma}; schema selection helps weaker text-to-SQL models in particular \citep{maamari2024death}. Orchestration splits the work among agents with their own contexts \citep{wu2023autogen,hadfield2025multiagent} and helps most when the single agent is weak \citep{kim2025towards}. Compaction bounds the history and, with optimized or trained compression, keeps accuracy at lower peak token use \citep{packer2023memgpt,zhou2026mem1,kang2025acon}. The same studies report limits, mostly for strong models: schema selection and a static semantic layer can lower the accuracy of capable models \citep{maamari2024death,chong2026evoontology}, a single agent given the same budget can match a multi-agent system \citep{tran2026single}, and plain summaries are no better than dropping old observations \citep{lindenbauer2025complexity}. The closest studies relate orchestration's benefit to the single agent's strength \citep{kim2025towards} and compare ways of delivering a semantic layer \citep{chong2026evoontology}; both judge by accuracy. We study the CETs together across seven LLMs and describe each by the behavior it changes and the failures it repairs, relative to the agent's habits.

\section{Practical Guidance and Conclusion}
\label{sec:guidance}

Context-engineering techniques are usually judged by their average effect; our results show that what they do depends on the agent's habits. The same CET can rescue one agent and hurt another, and it can hurt while the measures it was built to improve, tool use, calls, prompt size and the benchmarks' process scores, look better. The agents most at risk are the strongest: given a semantic layer, investigators treat it as the answer and stop checking, and under orchestration they hedge less and approve wrong answers. What helps weaker agents is not more compute but the CET that matches their habit: orchestration, which demands evidence, for agents that answer without looking, and a semantic layer for agents that give up. Compaction keeps prompts small but not the bill, and is safest with a layer, which keeps compacted runs from looping. None of the three fixes the most common remaining failure, choosing the wrong data after looking, which is where work on data agents should turn next.

\section*{Limitations}

Our study covers seven LLMs from four providers, from frontier models to a small open-weight model served locally, under all eight combinations of three CETs on five benchmarks. Models change quickly, and newer ones may show behaviors beyond the three groups we observe; the behavioral measures we use make it straightforward to place a new model among them, and extending the study to more models and model families is a natural next step. We study three widely used CETs, each in one representative design. Context engineering also includes other techniques, and other designs of the ones we study. Our aim is not to cover them all but to show that CETs interact with the agent's behavior and with each other in ways their average effects do not reveal; applying the same behavioral lens to other techniques and designs is promising future work. Our tasks ask agents to answer questions over heterogeneous and messy data from files, databases, documents and the web. Other domains, such as mathematical reasoning or software engineering, place different demands on agents, and whether the same behavior groups and interactions appear there is an interesting question for future work. Finally, we evaluate CETs as fixed configurations; systems that choose or adapt CETs to an agent's observed behavior, which our results suggest could pay off, are a direction for future work.

\section*{Ethical Considerations}
Our study evaluates context-engineering techniques for data agents, not their readiness for autonomous deployment. We find that the same technique can improve one model while degrading another, and that reduced cost or increased tool use does not necessarily imply more accurate answers. Semantic layers can encourage unsupported conclusions, compaction can omit relevant evidence, and orchestration can propagate errors or introduce unnecessary computation. These failures could have serious consequences when analytical outputs inform medical, legal, financial, or other high-stakes decisions. Such applications require task-specific validation, traceable evidence, and appropriate human oversight. Agents accessing files, databases, or web sources should also operate with least-privilege permissions and safeguards against unintended disclosure or modification of data. Finally, our behavioral groupings describe patterns observed under the evaluated tasks and configurations, rather than fixed properties of models. We encourage practitioners to validate context-engineering choices against answer correctness and evidence provenance, rather than relying on aggregate benchmark gains or efficiency measures alone.  

\section*{AI Use Statement} We used generative AI tools for assistance with coding and with editing the writing; all AI-assisted code was checked and tested by the authors, and all AI-assisted text was reviewed and revised by the authors. We did not use generative AI tools for research ideation, literature search or the analysis and interpretation of results. LLMs are also the objects of this study and serve as annotators and judges, as described in Sections 2-3 and the appendices. The authors take full responsibility for the content of this paper.

\bibliography{custom}

\clearpage
\appendix
\crefalias{section}{appendix}
\raggedbottom

\renewcommand{\topfraction}{0.9}
\renewcommand{\dbltopfraction}{0.9}
\renewcommand{\bottomfraction}{0.8}
\renewcommand{\textfraction}{0.07}
\renewcommand{\floatpagefraction}{0.75}
\renewcommand{\dblfloatpagefraction}{0.75}
\setcounter{topnumber}{3}
\setcounter{dbltopnumber}{3}
\setcounter{totalnumber}{5}

\noindent The appendices follow the order of the paper. \Cref{app:config,app:behavior,app:effort} give the setup: the agent configurations, the behavior measures and task demands, and reasoning effort. \Cref{app:fail,app:demand-profile,app:permodel} show how baseline agents fail, which tasks and agents each CET helps, and each CET's full behavioral fingerprint. \Cref{app:orch,app:layer,app:compaction,app:combine} give further results for orchestration, the semantic layer, compaction and their combinations, and \Cref{app:cost,app:published,app:compute} report cost, each benchmark's own metrics, and implementation and compute.

\section{Agent Configurations}
\label{app:config}

All eight configurations run in one harness with the same tools and decoding settings, and a budget of up to 300 steps and 600 model calls per task. With the semantic layer, the entries most relevant to the question, together with the join paths that connect them, are placed in the prompt up to a fixed size, and the agent can retrieve again by writing a new query. With orchestration, a finalizer writes the answer once the verifier has accepted the specialists' results. With compaction, each rewrite of the working memory is one extra model call, and the layer's entries sit outside the memory, so a rewrite never drops them. We use the FDABench report and multiple-choice tasks that have reference answers. \Cref{app:compute} gives the software, hardware and time per run.

\section{Behavior and Task Features}
\label{app:behavior}

\begin{figure*}[tp]
  \centering
  \includegraphics[width=\textwidth]{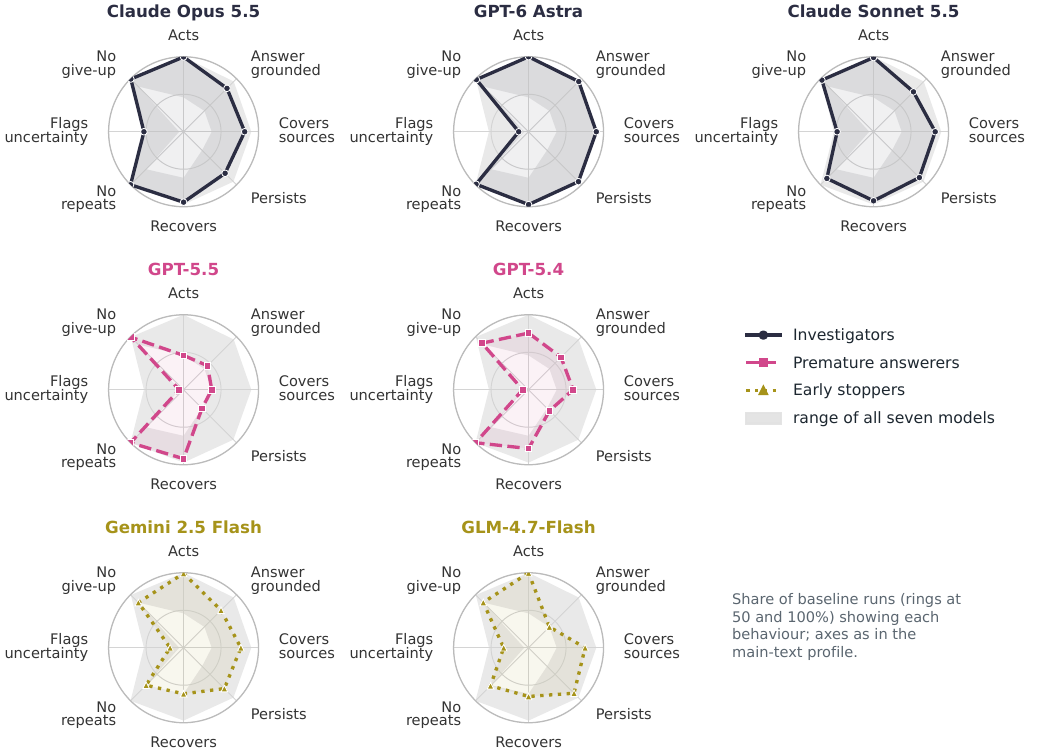}
  \caption{\textbf{Baseline behavior profile of every model.} Same axes as \Cref{fig:profile}, one radar per model, grouped by behavior group.}
  \label{fig:a1d}
\end{figure*}

\begin{figure*}[tp]
  \centering
  \includegraphics[width=\textwidth]{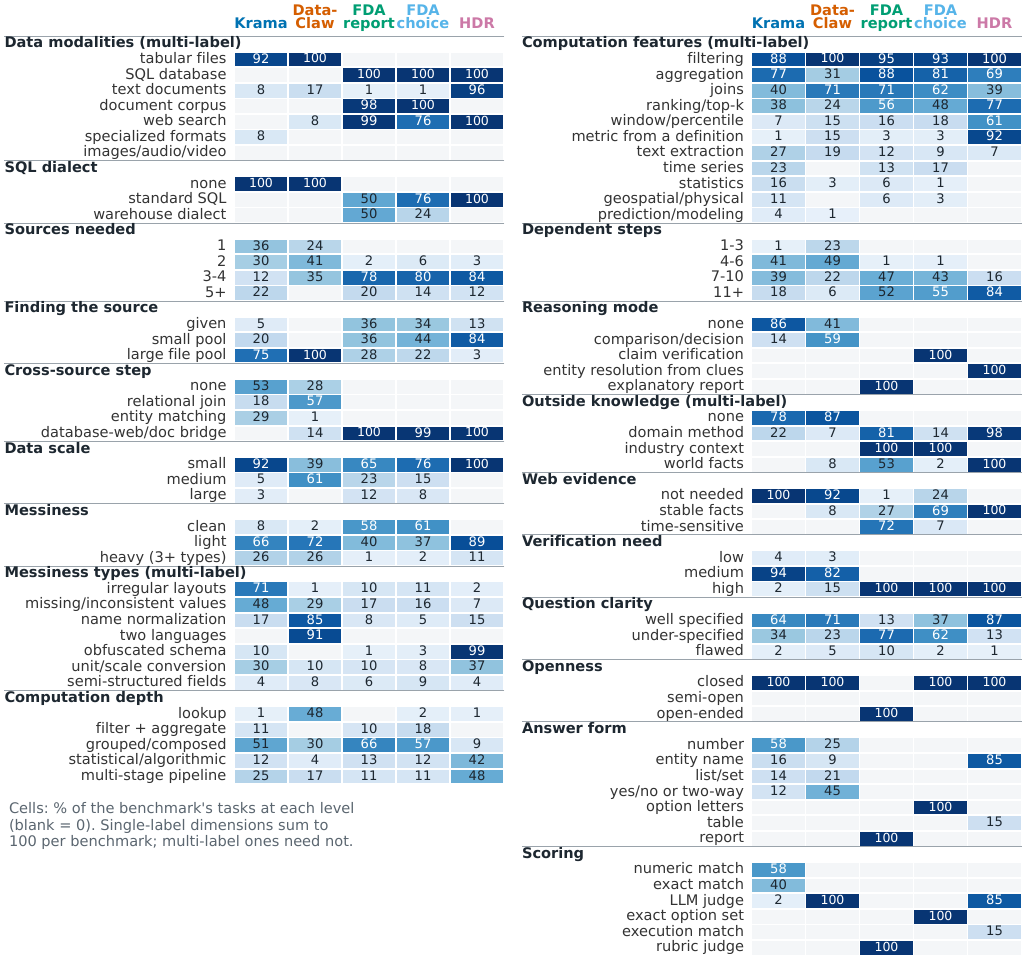}
  \caption{\textbf{Full demand profiles.} Share of each benchmark's tasks at each of the 88 levels of the 19 demand dimensions.}
  \label{fig:demands-full}
\end{figure*}

This section distinguishes the measurements used for clustering from those used to describe the resulting groups. Twelve trace features define the model groups and run types; eight behavioral indicators summarize their behavior in the radar charts. We also describe the annotation of uncertainty and stop reason and the 19 task-demand dimensions. \Cref{fig:a1d} shows each model's baseline behavioral indicators, and \Cref{fig:demands-full} shows each benchmark's task-demand profile.

\begin{figure*}[tp]
  \centering
  \includegraphics[width=\textwidth]{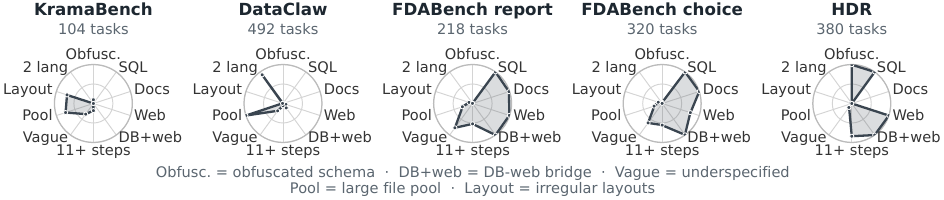}
  \caption{\textbf{Each benchmark demands something different.} Share of each benchmark's tasks that need each capability, with the same labels for every benchmark.}
  \label{fig:demands}
\end{figure*}

\paragraph{Trace features used for clustering.}
We normalize every trace to model steps and executed actions. An executed action is a call whose action block (Python, SQL, schema linking, web, vector or file operation) has a saved tool observation. For orchestrated runs, we count planner decisions and every specialist execution. We compute twelve features per run: whether the run acted, the share of steps with an executed action, the number of source families used, the number of distinct sources touched, the numbers of inspection actions, web searches and executions, the error rate of executed actions, the number of identical repeated actions, the number of actions after the last new information, whether the run ended by giving up, and whether it ended in another non-answer (budget, empty reply, refusal, abstention, plan). Counts are on a log scale.

\paragraph{Clustering models and individual runs.}
We distinguish model-level \emph{behavior groups} from run-level \emph{behavior types}. To identify model groups, we average the twelve standardized features over each model's baseline runs. We apply Ward clustering to these seven model feature vectors and partition them into three groups. Complete linkage yields the same groups; average and single linkage merge GPT-5.4 with the investigators or isolate it.

To identify individual run types, we apply $k$-means to the baseline runs using the same twelve features, standardized on baseline runs. We select the number of clusters by bootstrap stability measured with the adjusted Rand index. We freeze this baseline fit and assign runs from every configuration to the nearest baseline centroid, making run types comparable across CETs.

\paragraph{Behavioral indicators displayed in the radar charts.}
The radar charts use eight indicators to describe and interpret the behavior groups. Clustering uses the twelve trace features above; these indicators provide an interpretable summary of the resulting profiles.
\emph{Acts}: at least one executed action.
\emph{Answer grounded}: the submitted value appears in a tool output (KramaBench, DataClaw, HDR).
\emph{Covers sources}: mean share of the needed source types touched.
\emph{Persists}: three or more executed actions.
\emph{Recovers}: mean share of failed calls followed by a successful one.
\emph{No repeats}: no identical repeated call.
\emph{Flags uncertainty}: an annotator labels the answer as hedged.
\emph{No give-up}: the run does not stop with ``no data found''.

\paragraph{How the indicators distinguish models.}
In \Cref{fig:a1d}, the investigators have high values on most behavioral indicators; uncertainty flagging varies, with Opus flagging uncertainty more often than Astra. GPT-5.5 has low values for acting, answer grounding, source coverage and persistence. GPT-5.4 lies between GPT-5.5 and the investigators on these indicators, consistent with its less stable group assignment. Gemini and GLM act about as often as the investigators but give up and repeat actions more often, especially GLM.

\paragraph{LLM annotation of behavior.} GPT-5.4 (temperature 0, no reasoning, a fixed JSON schema) labels expressed certainty and stop reason for about 14,000 runs: all baseline runs, and 75 tasks per model in every configuration. It sees each run's steps and final reply but not the gold answer or the score. A second annotator, GPT-6 Astra, labeled 387 of these runs; agreement is high for certainty ($\kappa=.85$) and stop reason ($\kappa=.83$). Certainty and stop reason are the only behavior labels used in the main text.

\paragraph{Acting and accuracy.} Within benchmarks where GPT-5.5 sometimes acts, its acting runs score more than twice as high as its non-acting runs (.46 against .19 on KramaBench, .22 against .09 on HDR). Across all seven models, baseline answers given without any tool are right in 6.7\% of 1,247 runs, against 43.5\% of the runs that act. These runs are self-selected, so we read this as consistent with a habit rather than as proof of one; the orchestration results in \Cref{sec:orch} and the reasoning-effort runs in \Cref{app:effort} are the stronger evidence.

\paragraph{How many runs identify a group.} Drawing $n$ baseline tasks per model at random (1,000 draws, benchmarks in the model's own proportions) and assigning the nearest group by the share of runs without a tool, the give-up rate and the share of grounded answers recovers the model's group in 73\%, 82\%, 87\% and 92\% of draws at $n=5$, 10, 20 and 50. Astra, Opus and GPT-5.5 are recognized in 91-99\% of draws from ten runs, GLM in 95\% from twenty and Gemini in 90\% from fifty. GPT-5.4, which lies between the premature answerers and the others, is recognized in only 39-55\%.

\paragraph{Task demands.}
\label{app:demands}
We describe every task on 19 dimensions shared by all benchmarks. Two LLMs labeled all tasks and a third broke ties (mean Cohen's $\kappa=.86$ between the two over the single-label dimensions; 96\% of 330 hand-checked labels agree). The benchmarks differ widely on these dimensions (\Cref{fig:demands}), so we compare tasks with and without a demand within each benchmark. Six describe the data: the modalities involved (tabular files, SQL database, text documents, document corpus, web search, specialized formats, media), the SQL dialect, the number of sources needed, how the source must be found (given, small pool, large file pool), the cross-source step (none, relational join, entity matching, database-web or document bridge) and the data scale. Two describe messiness: its overall level, and its types (irregular layouts, missing or inconsistent values, name normalization, two languages, obfuscated schema, unit or scale conversion, semi-structured fields). Five describe the work: computation depth (lookup to multi-stage pipeline), computation features (for example joins, ranking, window functions, metrics defined in the task), the number of dependent steps, the reasoning mode, and the outside knowledge required. The remaining six describe the question and its evaluation: whether web evidence is needed and how recent it must be, the need to verify, question clarity, openness, answer form and scoring. \Cref{fig:demands-full} gives the share of each benchmark's tasks at every level; it is the detailed version of \Cref{fig:demands}, and the levels compared in \Cref{app:demand-profile} are taken from it.

\begin{figure*}[tp]
  \centering
  \includegraphics[width=\textwidth]{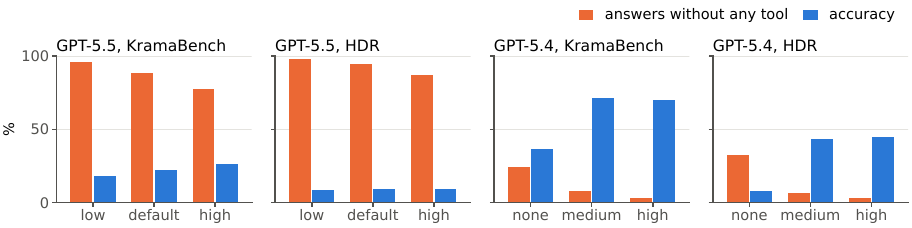}
  \caption{\textbf{Reasoning removes GPT-5.4's habit of answering before looking, but not GPT-5.5's.} Share of baseline runs that answer without any tool (orange) and accuracy (blue) at each reasoning effort, on KramaBench and HDR. GPT-5.4 does not reason at its default; GPT-5.5 does.}
  \label{fig:effort}
\end{figure*}

\section{Reasoning Effort}
\label{app:effort}

All main runs use provider defaults, under which every model reasons except GPT-5.4 (the Claude models use adaptive thinking, which Claude Opus 5.5 cannot turn off; \citealp{anthropic2026opus55,anthropic2026sonnet55}). To see whether answering before looking is a setting or a habit, we ran the baseline agent of both models on KramaBench (104 tasks) and HDR (380 tasks) at other reasoning efforts: low and high for GPT-5.5, medium and high for GPT-5.4 (\Cref{fig:effort}). Everything else is unchanged. In the figure, the orange bars fall and the blue bars rise for GPT-5.4 as effort grows, while both stay nearly flat for GPT-5.5.

For GPT-5.4, reasoning removes the habit. Its share of tasks answered without a tool falls from 24\% to 8\% (medium) and 3\% (high) on KramaBench, and from 32\% to 7\% and 3\% on HDR. Its accuracy rises by 35 points on KramaBench (.37 to .72 and .70; +35.1 [+25.1, +44.6] at medium, 57 tasks better and 8 worse) and by 35-36 points on HDR (.08 to .43 and .45; 157-162 tasks better, 16-17 worse). Both values exceed the best of its eight configurations without reasoning (.63 and .23). The cost per task rises from \$0.07 to \$0.18-\$0.21 on KramaBench and from \$0.10 to \$0.53-\$0.66 on HDR, about what orchestration costs it.

For GPT-5.5, more reasoning moves the habit only a little. At high effort it still answers without a tool on 78\% of KramaBench and 87\% of HDR tasks (89\% and 94\% at the default), and its accuracy does not change clearly (+4.3 [$-$2.1, +11.3] and +0.2 [$-$2.4, +2.9] points). Low effort raises the share without a tool to 96\% and 98\%. When it does act at high effort, its accuracy is again more than twice as high (.49 against .20 on KramaBench). Answering before looking therefore has two forms, with two different remedies. In GPT-5.4 it comes with reasoning switched off, and switching reasoning on gives a larger gain than orchestration at about the same cost. In GPT-5.5 it persists at every effort: more thinking does not make the model look, and only orchestration, which asks for evidence from outside the model, does (\Cref{sec:orch}). Reasoning and orchestration are therefore not interchangeable ways of spending more compute; which one helps depends on why the agent does not look.

\section{How Baseline Agents Fail}
\label{app:fail}
\label{app:annotation}

\begin{figure*}[tp]
  \centering
  \includegraphics[width=\textwidth]{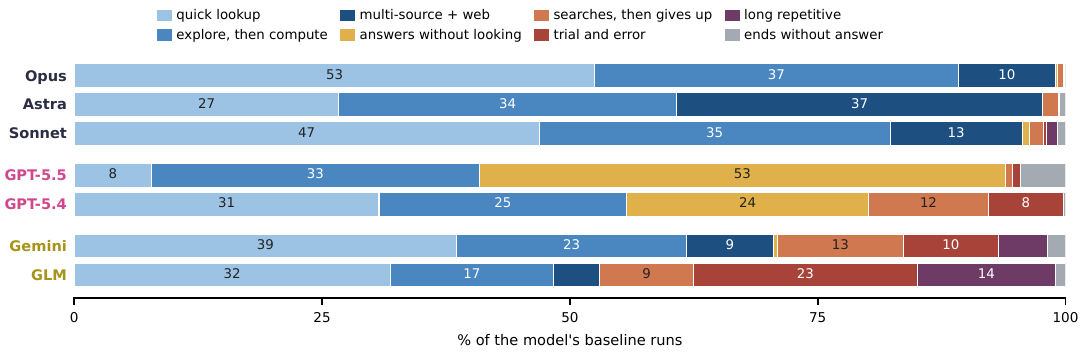}
  \caption{\textbf{Behavior types by model.} Share of each model's baseline runs in eight behavior types. Investigators spend almost all runs in three productive types; GPT-5.5 answers without looking in half of its runs; GLM spends a third of its runs in trial and error and repetitive loops.}
  \label{fig:types}
\end{figure*}

\begin{figure*}[tp]
  \centering
  \includegraphics[width=\textwidth]{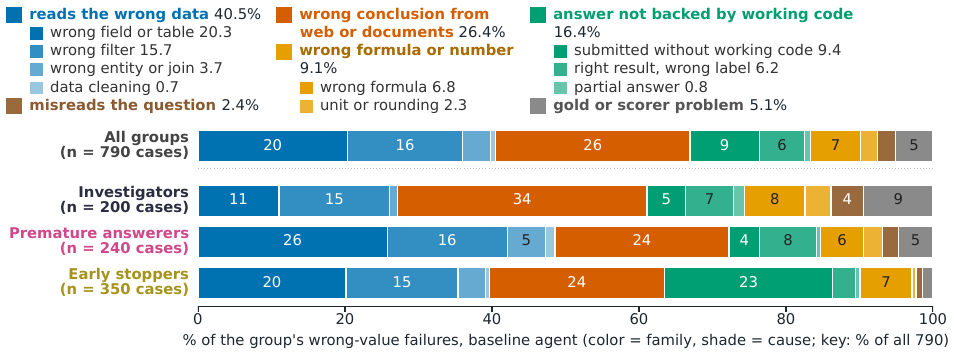}
  \caption{\textbf{Most wrong values after looking come from the wrong slice of data or from external evidence, not from arithmetic.} Share of each group's wrong values in twelve causes grouped into six families (color: family; shade: cause), from 790 annotated baseline wrong values weighted to the wrong values they represent.}
  \label{fig:causes}
\end{figure*}

This section describes how baseline runs end and why the wrong ones are wrong. We first give the setup: how runs are sorted into behavior types and failure types, and how wrong values were annotated with causes. We then give the results: the behavior types of each model (\Cref{fig:types}), its failure rate and failure types, and the causes of wrong values by group (\Cref{fig:causes}), including how the CETs shift them. We end with how far the annotators agree.

\paragraph{Setup.} Behavior types are the $k$-means types of \Cref{app:behavior}. Failure types come from the trace and the score: a run fails with a wrong value after looking, by using only some of the needed sources, by answering without any tool, by giving up, or otherwise. For wrong values, three LLM annotators (GPT-5.5, GPT-6 Astra and GPT-5) assign each case one of twelve causes in six families from the run's trace, using one codebook and prompt. We use the majority label; GPT-5.5 adjudicates the 123 cases (8.5\%) without a majority. A fourth rater (Claude Opus 5.5) serves as a vendor-independent check. Cases are sampled within each behavior group over model-by-benchmark cells in proportion to the failures in each cell, with at least two per cell, and weighted by the number of failures each case represents. The sample has 790 baseline wrong values (200 investigators, 240 premature answerers, 350 early stoppers) and 220 each with orchestration, with the semantic layer and with compaction (70, 90 and 60 per group).

\paragraph{Investigators spend almost all their runs in productive types.} The investigators spend almost all their runs in three productive types, quick lookups, exploring then computing, and multi-source and web investigations (\Cref{fig:types}). GPT-5.5 answers without looking in 53\% of its runs and GPT-5.4 in 24\%. The early stoppers spread over more types: GLM spends 23\% of its runs in trial and error and 14\% in long repetitive runs, and both models search and then give up in 9-13\%.

\paragraph{Early stoppers fail most, but not for lack of effort or data.} The same split shows in the failures. The investigators fail in 28-38\% of their baseline runs, mostly with a wrong value after looking (18-20 points) or by using only some sources (7-15). GPT-5.5 fails in 70\%, two thirds of them because it answered with no tool (47 points); GPT-5.4 fails in 81\%, spread over wrong values, missing sources, no tool and giving up. The early stoppers fail in 81-87\%, most often by using only some of the sources (37-42 points), and give up in 14-15\%. Their give-ups are not for lack of effort or data: in 445 baseline give-up runs (427 of them on DataClaw) they had touched every kind of source the task needs in 99\% of runs, after a median of 16 executed actions against 5 in an investigator's run, and at least one investigator solves 80\% of these tasks. Asked for a region's R\&D intensity, for example, Opus took a stored average instead of computing it from totals, although its own alternative calculation had shown the right value.

\paragraph{Most wrong values come from the wrong data slice or from external evidence.} Most wrong values after looking come from the wrong slice of data or from external evidence, not from arithmetic (\Cref{fig:causes}). With this sample the four main families have 95\% intervals of about $\pm3$ points: wrong data slice 40.5\% [37.2, 43.8], external evidence 26.4\% [24.1, 28.9], answer production 16.4\% [13.9, 19.0], computation 9.1\% [7.1, 11.2]. Investigators hold the right value in their own tool output in 32\% of their wrong values against 13\% for premature answerers (+19.5 points [+11.9, +27.1]; every rater gives +16 to +24); a rule-based count over all baseline wrong values gives 25\% against 10\%. Investigators select the wrong data less often than premature answerers ($-$21.4 points [$-$29.1, $-$13.5]), early stoppers fail in producing the answer more often than the other two groups (+13.9 [+8.4, +19.5]), and investigators draw wrong conclusions from external evidence more often than premature answerers (+10.3 [+4.3, +16.4]).

\paragraph{No CET removes the main causes of wrong values.} Against baseline runs, orchestration lowers the share of answer-production failures ($-$7.5 [$-$11.8, $-$3.1]), mostly through code that never ran (9.4\% to 3.9\%), and so does the semantic layer ($-$7.1 [$-$11.4, $-$2.4]; early stoppers $-$15.3 [$-$23.9, $-$6.4]); compaction leaves every family within $\pm$2.4 points. Wrong filters do not rise with orchestration (15.7\% to 18.6\%, +2.9 [$-$2.5, +8.6]). The causes are therefore much the same across groups and CETs. Although the three groups behave very differently, in every group choosing the wrong data or drawing the wrong conclusion from documents or the web accounts for most wrong values (61\% for the investigators, 72\% for the premature answerers, 64\% for the early stoppers) and computation for only 8-12\%; the groups differ only in which of the two leads, and the CETs move the failures to produce working code but not these two. Once an agent looks, it goes wrong in the same ways, whatever its habits and whichever CET it uses.

\paragraph{Agreement between annotators.} Agreement on the primary cause is moderate (Fleiss $\kappa=.60$ for the three raters, .63 with the fourth; pairwise .54-.68), and most disagreements cross families (wrong field against wrong filter, misread question against wrong filter). Single raters differ by up to 14 points on the share of wrong data slices (32-46\%), so we report the majority vote and read single causes with care. A fourth rater from another vendor (Opus) changes no family share by more than 2 points.

\FloatBarrier
\section{Where and for Whom Each CET Helps}
\label{app:demand-profile}

\begin{figure*}[tp]
  \centering
  \includegraphics[width=\textwidth]{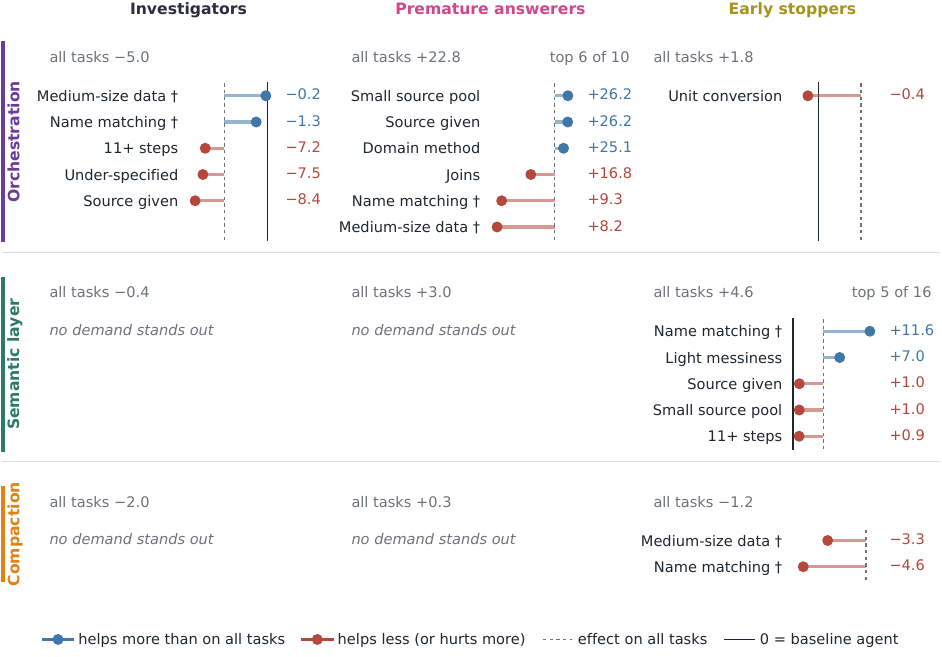}
  \caption{\textbf{Each CET changes size on a few demands, and only for some groups.} For each CET (rows) and behavior group (columns), the demands on which the effect against the baseline agent departs from the group's effect on all tasks (dashed line) by at least 2 points, with a 95\% task-bootstrap interval that excludes it; at most the three largest departures in each direction (``top $k$ of $n$'' when more qualify). Blue: helps more than on all tasks; red: helps less or hurts more. Values are points of accuracy, groups are the mean of their models, and each cell has its own scale. $\dagger$: at least 74\% of the level's tasks come from DataClaw.}
  \label{fig:demand-profile}
\end{figure*}

This section asks which tasks and which agents each CET helps. It first breaks each CET's effect down by task demand (\Cref{fig:demand-profile}), then asks what predicts the size of an effect across models and benchmarks.

\begin{figure*}[tp]
  \centering
  \includegraphics[width=\textwidth]{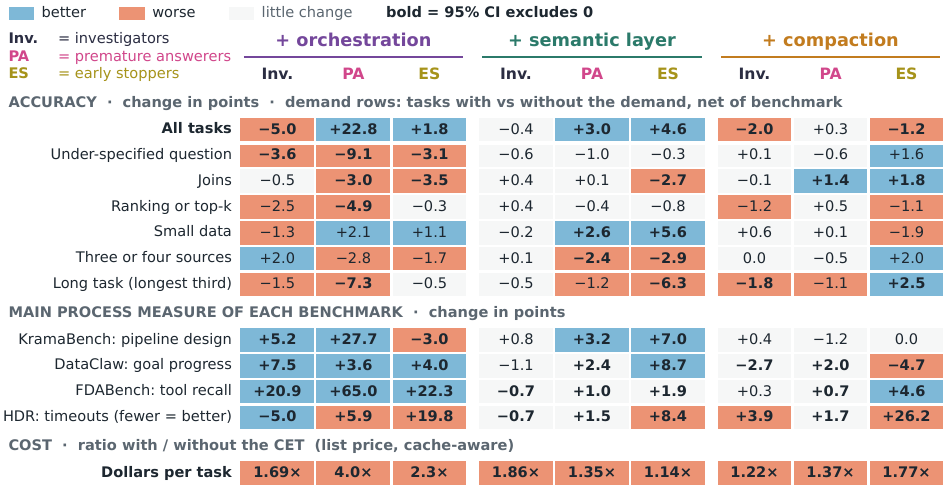}
  \caption{\textbf{Every demand and every benchmark's process measure.} Same as \Cref{fig:scorecard}, with three more demands and each benchmark's process measure in place of their mean. HDR has no step score, so its row is the share of runs that time out (fewer is better).}
  \label{fig:scorecard-full}
\end{figure*}

\paragraph{The group sets the sign of an effect, and the demand changes its size.} \Cref{fig:demand-profile} lists, for each CET and group, the demands on which the effect departs from the group's effect on all tasks, among the 19 demand levels that occur in at least two benchmarks (\Cref{app:demands} defines them). The group sets the sign of the effect, and the demand at most changes its size: orchestration lowers the investigators' accuracy on 18 of the 19 levels and raises the premature answerers' on all 19. Four of the nine CET-group pairs have no standout demand. Orchestration has the clearest profile: for the premature answerers it helps most when the source is given or the pool of candidate sources is small (+26 points) and least on joins (+17), name matching (+9) and medium-size data (+8); for the investigators it costs most on under-specified questions, long tasks and when the source is given ($-$7 to $-$8); for the early stoppers its small gain disappears on unit conversion ($-$0.4). The layer's profile belongs to the early stoppers: it helps them most where names must be matched (+11.6) and on lightly messy data (+7.0), and only about one point when the source is given, the pool is small or the task is long, that is, where the difficulty is reading the data rather than finding it. Compaction stands out only for the early stoppers, who lose most where names must be matched ($-$4.6) and on medium-size data ($-$3.3). Name matching and medium-size data come mostly from DataClaw, so these rows partly reflect that benchmark.

\paragraph{Process scores rise across demands, even where accuracy falls.} \Cref{fig:scorecard-full} extends \Cref{fig:scorecard} to six task demands and to each benchmark's own process measure, including the share of HDR runs that time out. On under-specified questions the orchestrated agents do not look less; they reason differently: the investigators stop stating the assumptions an ambiguous question requires (from 23\% of runs to 8\%). The other demands follow the same pattern. The weaker groups gain less from orchestration on tasks that need joins ($-$3.0 and $-$3.5 points) and the premature answerers on rankings ($-$4.9), where the early stoppers' specialists fail and loop more often. With the layer, the early stoppers give up and repeat themselves less on small data, and the weaker groups gain less when three or four sources must be combined ($-$2.4 and $-$2.9). On the longest third of tasks (those on which other agents run longest), compaction cuts the early stoppers' give-ups by 6 points and raises their accuracy, while the investigators repeat themselves, loop and run out of budget more often. On the process rows, the investigators already searched the web in 87\% of runs and queried the database in 93\%, so orchestration's 21-point rise in their FDABench tool recall does not reflect new kinds of source; DataClaw's goal progress and KramaBench's pipeline score also rise for them (+7.5 and +5.2 points) while their accuracy stays flat or falls.

\paragraph{The habit a CET repairs predicts its effect better than weakness.} Across the 35 model-benchmark pairs, all three CETs follow the same rule in sign: they help most where the baseline agent is weakest (Spearman correlation with baseline accuracy $-.65$ for orchestration, $-.43$ for the layer and $-.16$ for compaction). The habit a CET repairs predicts its effect better than weakness alone. Orchestration's gain rises with the baseline agent's share of answers without a tool ($+.60$): GPT-5.5 and Gemini have about the same baseline accuracy (.21 and .25 on the five-benchmark mean), yet GPT-5.5 gains 35 points from orchestration and Gemini less than one, because GPT-5.5 does not look and Gemini does. The layer's gain does not rise with the give-up rate across pairs ($+.12$), because give-ups are frequent only on some kinds of tasks; across the seven models, those that give up more gain more ($+.79$). Compaction's effect falls with the length of the baseline agent's runs ($-.48$): it costs most where the baseline agent already keeps a long, useful history. Weakness describes what the CETs do on average; the habit a CET targets tells which agent it is for.

\FloatBarrier
\section{Behavioral Fingerprint}
\label{app:permodel}

\begin{figure*}[tp]
  \centering
  \includegraphics[width=\textwidth]{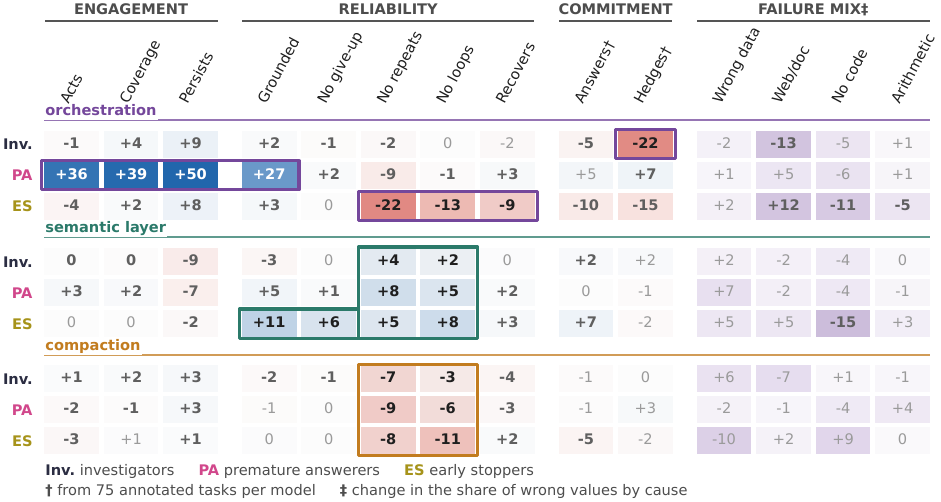}
  \caption{\textbf{Orchestration raises engagement, the layer raises reliability at the expense of engagement, and compaction lowers reliability.} Change in each behavior, in points of runs, from the agent without the CET to the agent with it, for each behavior group; blue is more of the property, red less, and bold an interval that excludes no change. Outlined cells are the main shifts; other cells are muted. The failure mix covers wrong data slice, wrong conclusion from web or document evidence, code that never ran and arithmetic, with purple for the size of the shift. } 
  \label{fig:fingerprint}
\end{figure*}

\begin{figure*}[tp]
  \centering
  \includegraphics[width=\textwidth]{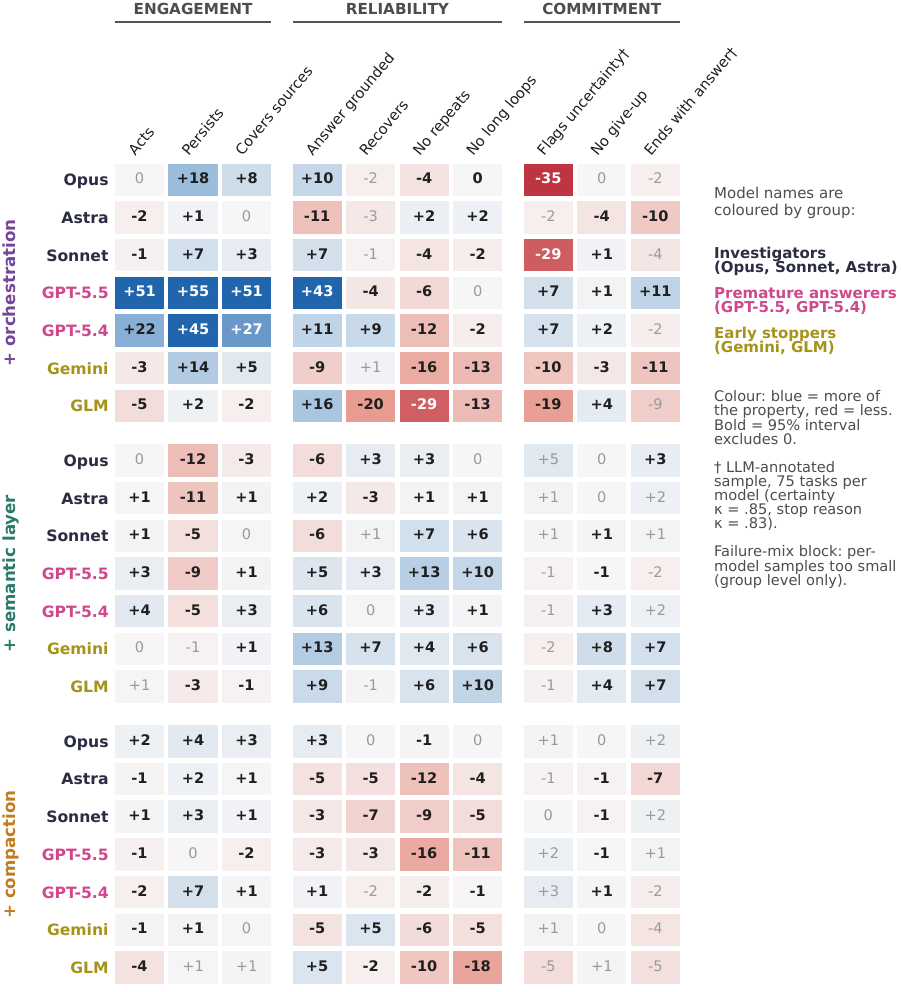}
  \caption{\textbf{Behavioral fingerprint by model.} Same as \Cref{fig:fingerprint}, one row per model.}
  \label{fig:fingerprint-models}
\end{figure*}

\Cref{fig:fingerprint} gives the full behavioral fingerprint behind \Cref{fig:shifts}: every measure, including loops, ending with an answer and the mix of failure causes, with the main shifts outlined. \Cref{fig:fingerprint-models} repeats it for every model. Most group shifts are carried by every member, so the groups are not the effect of one model. The exceptions are informative. Orchestration breaks the investigators in opposite ways: Opus and Sonnet flag uncertainty far less with orchestration ($-$35 and $-$29 points), while Astra does not, and instead its verifier rejects usable evidence until the planner abstains (\Cref{sec:orch}); orchestration makes the Claude models overconfident and Astra overcautious. It also builds the most investigation where there was least: GPT-5.5 gains more than GPT-5.4 in acting, persisting and covering sources (+51 to +55 against +22 to +45). With the layer, the drop in persistence comes from Opus and Astra ($-$12 and $-$11), the models that stop looking when the data is described. With compaction, the largest rise in repeated actions is spread across the groups (GPT-5.5 $-$16, Astra $-$12, GLM $-$10 on no repeats), as expected if loops follow the task rather than the model's habits (\Cref{app:compaction}).

\section{Orchestration: Further Results}
\label{app:orch}

\begin{figure*}[tp]
  \centering
  \includegraphics[width=\textwidth]{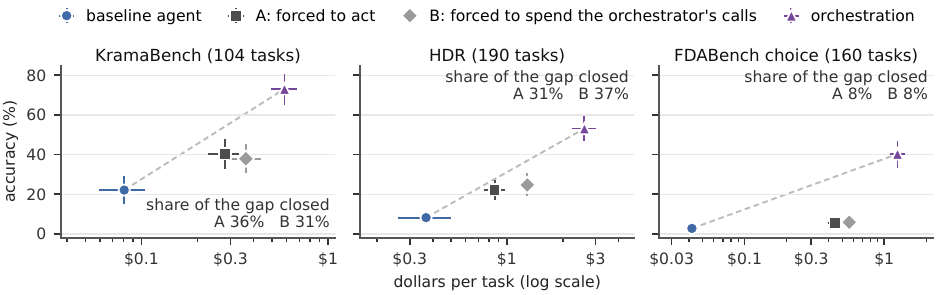}
  \caption{\textbf{Forcing a single GPT-5.5 agent to act, or to spend the orchestrated agent's calls, closes only 7-37\% of orchestration's gain.} Accuracy against dollars per task for GPT-5.5: the baseline agent, the single agent forced to act before answering (A), the single agent forced to spend the orchestrated agent's number of calls on each task (B) and the orchestrated agent; 95\% intervals over tasks.}
  \label{fig:compute}
\end{figure*}

\begin{figure*}[tp]
  \centering
      \includegraphics[width=\textwidth]{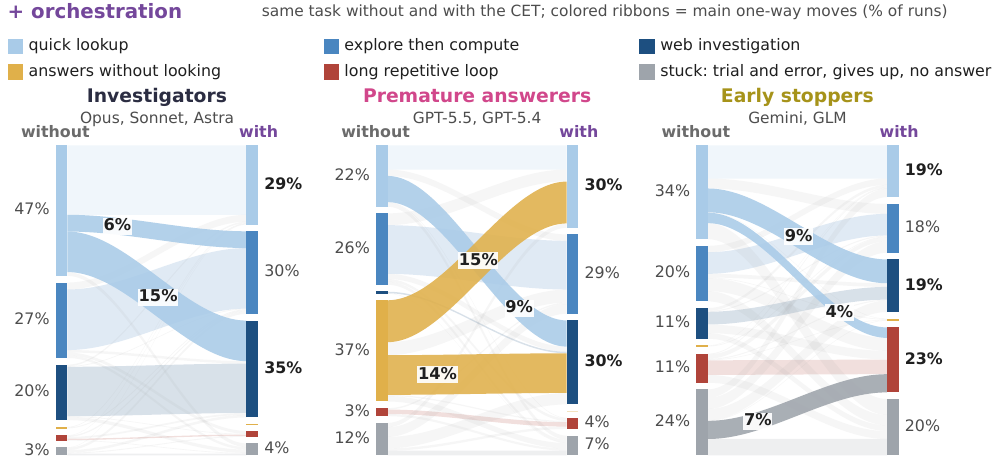}
  \caption{\textbf{Orchestration moves premature answerers' runs into looking.} Behavior types of runs on the same tasks without (left) and with orchestration (right), for each behavior group; bold ribbons are the main moves.}
  \label{fig:flow}
\end{figure*}

\begin{figure}[tp]
  \centering
  \includegraphics[width=\columnwidth]{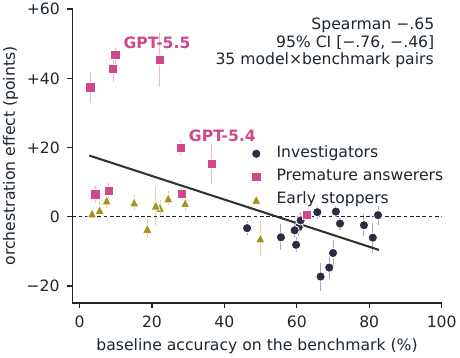}
  \caption{\textbf{Orchestration helps most where the baseline agent is weakest.} Accuracy change of orchestration against the baseline agent's accuracy on the benchmark, one point per model and benchmark (Spearman $-.65$; against the baseline share of answers without a tool, $+.60$).}
  \label{fig:scatter}
\end{figure}

This section backs the orchestration results of \Cref{sec:orch}. It shows that the gain is not a larger budget (\Cref{fig:compute}), how orchestration changes the kind of run (\Cref{fig:flow}), where accuracy is lost and won, and how the gain depends on the baseline agent's accuracy (\Cref{fig:scatter}).

\paragraph{Orchestration's gain is not a larger budget.} Orchestration spends more than the baseline agent: for GPT-5.5 it makes 4 to 16 times as many model calls per task. Its gain could therefore be a budget effect, or simply the effect of making a model that answers before looking take one look. We separate the two with two single-agent arms for GPT-5.5, the model with the largest orchestration gain, on KramaBench, HDR and FDABench multiple-choice. In arm A, the harness rejects any final answer given before the agent has run an action. In arm B, it also rejects any final answer until the agent has used as many model calls as the orchestrated agent used on the same task. Everything else is the baseline agent's setup, and a neutral notice explains each rejection. Both arms move only a short way from the baseline agent toward the orchestrated agent (\Cref{fig:compute}). On KramaBench, A and B raise accuracy from .22 to .40 and .38, against .73 for the orchestrated agent; on HDR from .08 to .22 and .25, against .53; on FDABench multiple-choice from .03 to .06, against .40. They close 31-37\% of the gap on KramaBench and HDR and 7-8\% on FDABench multiple-choice. B is not clearly better than A on any benchmark, although it used as many calls as the orchestrated agent on every task.

\paragraph{Forced agents take one look, then repeat their answer.} The transcripts show why: the agent runs one action (1.0 to 1.2 per run on average), then repeats an answer to every notice until the budget is spent, and in 50-86\% of runs that answer is the one it gave before looking. Forcing moves GPT-5.5 from answering without looking to a quick lookup, never to the multi-step, multi-source work of orchestrated runs; on HDR, 5\% of forced runs use the web, against 98\% of orchestrated runs. The forced arms cost 33-62\% of the orchestrated agent's dollars per task, so they are cheaper but not a substitute. Calls made available are not calls used: what orchestration adds is a planner that asks for specific evidence and specialists that must return it, a structure that this model does not supply for itself even when the budget is paid for. A gate that forces one look is a cheap partial repair where a single lookup suffices, as on KramaBench, and almost none where the answer needs several steps, as on FDABench multiple-choice. 

\paragraph{Orchestration builds an investigation only where there was none.} The flow in \Cref{fig:flow} shows the main moves: the premature answerers' runs that answered without looking become quick lookups (15\% of runs) and web investigations (14\%); the investigators' quick lookups turn into web investigations (15\%) and explorations (6\%); and the early stoppers' long repetitive runs double (11\% to 23\%). For the investigators, orchestration changes little in how they work: the tool mix stays the same, they touch about two more files or tables and make 2.3 more calls per task, and the main change is that their replies hedge less. For the premature answerers, it builds an investigation where there was none: runs that answer without a tool fall by 42 points, and SQL, web search and document search each enter about half of the runs (+51 to +62 points) at 14 more calls per task. For the early stoppers, it adds effort without adding reach: 39 more calls, about 4,700 more characters per FDABench report and more repeated actions, but no additional kinds of source.

\paragraph{Orchestration is three different interventions.} Splitting each group's change in accuracy into points lost (baseline right, orchestrated wrong) and won (the reverse), by cause, shows what these moves do. The premature answerers win 27.5 points, 15.3 of them on tasks their baseline agent answered without any tool. The investigators lose 11.4 points and win back 6.4; their largest losses are wrong values although every needed source was reached (4.1 points) and dropped or swapped options on multiple-choice claim checks (3.4), where recall of the correct options falls from .88 to .76 for Opus and from .87 to .72 for Astra. The loss is confined to tasks that chain a database with documents and web search; on tasks that stay inside files or a single database, the investigators lose nothing. The early stoppers lose and win across all causes in similar amounts (9.5 against 11.2). Orchestration is thus three different interventions: a change of tone for the investigators, a new way of working for the premature answerers, and more of the same for the early stoppers.

\paragraph{Orchestration pays where the baseline agent does not investigate.} Whether orchestration helps depends on how well the baseline agent already does. In \Cref{fig:scatter}, each point is one model on one benchmark: the lower the baseline agent's accuracy, the larger orchestration's gain (Spearman $-.65$, 95\% CI $[-.76, -.46]$, 35 model-benchmark pairs). Orchestration lowers accuracy in 13 of the 17 pairs where the baseline agent scores above .45 (mean $-$4.7 points) and raises it in 17 of the 18 pairs at or below .45 (mean +14.0), the threshold at which \citet{kim2025towards} also find that adding agents stops paying. Behavior predicts the gain almost as well as accuracy does: it correlates at $+.60$ with the share of tasks the baseline agent answers without any tool. The four largest gains (+37 to +47 points) are GPT-5.5's, three of them on benchmarks where its baseline agent answers 89-99\% of tasks without a tool; the investigators' 15 pairs all lie at the right, 12 of them below zero and none above +1.5 points. Before adding orchestration, it is therefore worth measuring how the baseline agent does and how often it answers without looking; orchestration pays where the baseline agent does not investigate, not where it investigates well.

\section{Semantic Layer: Further Results}
\label{app:layer}

\begin{figure*}[tp]
  \centering
  \includegraphics[width=\textwidth]{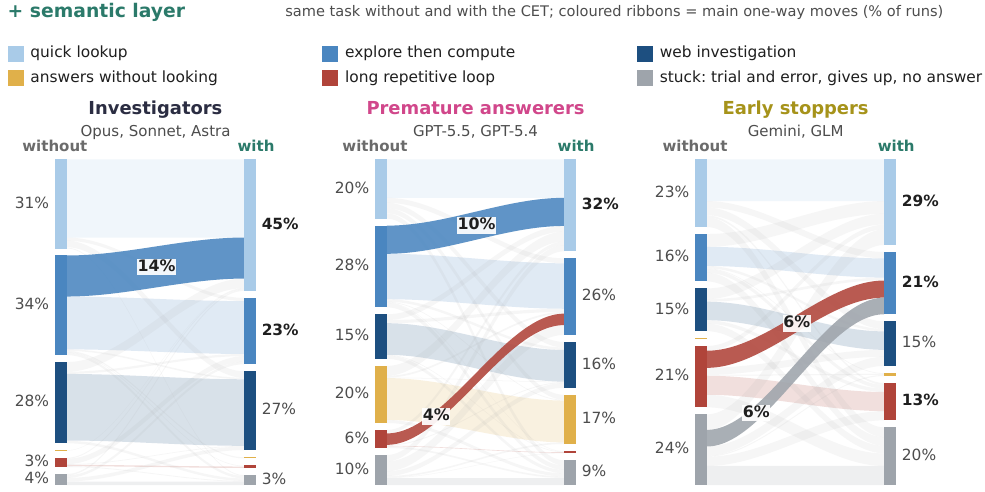}
  \caption{\textbf{The layer moves every group toward quick lookups: it replaces failure for early stoppers and work for investigators.} Behavior types of runs without and with the layer, by group.}
  \label{fig:layer-flow}
\end{figure*}

\begin{figure*}[tp]
  \centering
  \includegraphics[width=\textwidth]{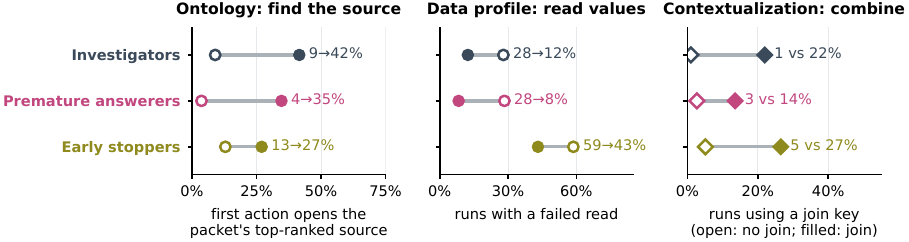}
  \caption{\textbf{Each part's content reaches the agent's actions where it is meant to.} Without (open) and with (filled) the layer: runs whose first action opens the source the layer lists first (KramaBench and DataClaw, both architectures); runs with a failed read (KramaBench, single agent); runs that use a join key from the layer on KramaBench tasks that need no join (open) or need one (filled). An element counts as used only when it appears in an executed action.}
  \label{fig:parts-profile}
\end{figure*}

\begin{figure*}[tp]
  \centering
  \begin{minipage}[b]{0.49\textwidth}\centering\includegraphics[width=\linewidth]{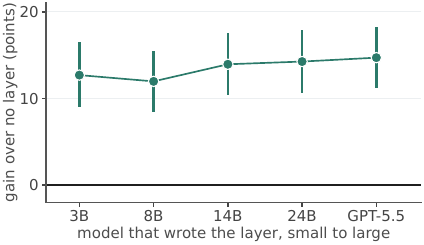}\end{minipage}\hfill
  \begin{minipage}[b]{0.49\textwidth}\centering\includegraphics[width=\linewidth]{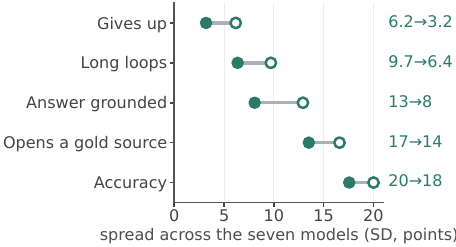}\end{minipage}
  \caption{\textbf{A small author suffices, and the layer makes models alike in how they work more than in accuracy.} Left: accuracy gain of a GPT-5.4 single agent over its no-layer run for layers written by models of increasing size on DataClaw (28\% accuracy without a layer), with 95\% intervals paired by task. Right: spread across the seven models (standard deviation, points) of each measure without (open) and with (filled) the layer; reaching a gold source is measured on single-agent KramaBench and DataClaw runs, the other measures over all runs.}
  \label{fig:author}
\end{figure*}

This section backs the layer results of \Cref{sec:layer}: how the layer changes the kind of run (\Cref{fig:layer-flow}), what each of its three parts does (\Cref{fig:parts-profile}), what the layer covers and what still fails, and how much its author matters and whether it makes the models alike (\Cref{fig:author}).

\paragraph{The layer moves every group toward quick lookups, for different reasons.} The layer moves every group toward a quick lookup, for different reasons (\Cref{fig:layer-flow}). For the early stoppers it replaces failure: their long repetitive runs fall from 21\% to 13\% of runs and their stuck runs from 24\% to 20\%. For the investigators it replaces work: 14\% of their runs move from exploring then computing to a quick lookup, and quick lookups rise from 31\% to 45\% of their runs. Whether the layer makes an agent stop or start looking depends on the model rather than its group: Opus stops looking in 73 runs and starts in 11, and Gemini in 68 against 26, while Sonnet, GPT-5.4 and GPT-5.5 start more often than they stop. The same layer that makes GPT-5.4 look makes Opus stop.

\paragraph{How we check each part.} Entries are retrieved through two channels, a lexical one that matches exact identifiers, codes and units and a dense one that matches the meaning of the information need, and the agent can retrieve again by writing a new query for each channel. The layer has three parts (\Cref{sec:framework}): the ontology helps the agent find the right source, the data profile helps it read values correctly, and the contextualization helps it combine sources. We check each part where it is meant to help, comparing runs with and without the layer on the same task rather than adding the parts one by one, and count a part as used only when its content shows up in the agent's code, SQL or tool calls. To see which part decided a changed answer, an LLM judge (GPT-5.4) read 600 such run pairs with the layer split into its three parts; a second judge (GPT-5.5) agrees moderately ($\kappa=.41$).

\paragraph{Each part does its job, and the ontology decides the most answers.} Each part does its job (\Cref{fig:parts-profile}). With the ontology, agents start from the right place more often: the first file they open is the one the layer lists first in 27-42\% of runs, against 4-13\% without it. With the data profile, fewer runs fail to read a file: on KramaBench the share falls from 28\% to 12\% for the investigators, from 28\% to 8\% for the premature answerers and from 59\% to 43\% for the early stoppers. With the contextualization, agents use join keys where a join is needed (14-27\% of runs) and almost never elsewhere (1-5\%). Of the three, the ontology decides the most answers: per 1,000 runs, 47 fixes and 17 breaks, against 6 and 2 for the data profile and at most 5 and 2 for the contextualization. It helps most the agents that search least: it brings the early stoppers to a source they had missed in 619 runs and away from one in 427, while for the investigators, who had usually found the source themselves, the balance is 77 against 448. Where a run applies a unit or format the layer states, wrong answers become right four times as often as the reverse; Opus, for example, divides an HDR success rate by 100 because the layer says it is a percentage, and answers right. Runs that use a join key from the layer score 20 points higher than runs that do not. Because the layer's entries ride in every prompt, input tokens rise 1.7 to 2.4 times.

\paragraph{Agents use the layer as a guide, not as an answer.} The layer is built to prevent the baseline agent's failures (\Cref{sec:plain}). It names several candidate files, and the right one is among them in 92-96\% of cases; it says what each column means, with units and example values; it says how to read each file and how files join; and it tells the agent to check everything with its tools. Agents use it as a guide, not as an answer. When the first file it names is the wrong one, they open that file first about as often as without the layer (9-23\% of runs, against 5-31\%), most go to other files instead, and only 0-3\% answer from it alone. What it gives them is hard to find on their own: for 80 of the 107 answers the ontology fixed, the judge found that the deciding fact would have taken further searching without the layer. Sometimes one line is enough: on a KramaBench legal task, GPT-5.4 skips the two header rows the data profile mentions and gets 5,435, the right answer, where without it it reads 6,438.

\paragraph{Breaks come from skipped checks more than from the layer.} When an answer still breaks, the layer is rarely the cause: in a sample of 300 broken answers, it misled the agent in 24 and played no part in 198. In the remaining 78, the agent stopped checking because the layer looked complete, chose a similar file over the right one, or opened the right file and read the wrong sheet or column. These breaks were easy to avoid: in 39 of the 64 that the ontology decided, the right information was one obvious step away. The investigators are not fooled by the layer so much as tempted to skip the check: the investigators were misled in 2 of 100 sampled breaks and stopped checking in 25, while the early stoppers were misled in 13. In most of the investigators' stop-checking breaks the layer was correct and described the common case: it named one file, column, key or grain, and the agent took that as the whole picture, where its baseline run had inspected the data and found the second column, key or source. On KramaBench's beach water-quality tasks, for example, the layer correctly says each datasheet holds one Enterococcus value per row; with it, Opus and Sonnet read only that column, where without it they had inspected the headers and also used the per-site columns. The layer can say what each column is, but not which one a given question needs; that choice, and the choice to check, stay with the agent. Its fixes show the same: most come from saying what a column means (48 of 107).

\paragraph{A small author suffices, and the layer makes models alike in how they work.} Writing a layer costs one pass of a model over every source, so it matters whether a small model can write it. On DataClaw, a layer written by a 3B model gives a GPT-5.4 single agent nearly the same gain as one written by GPT-5.5 (+12.7 against +14.7 points; \Cref{fig:author}, left), and every author's gain excludes zero. Because the layer helps the weaker models most, one may also expect it to make the models alike. It does so in how they work far more than in what they score (\Cref{fig:author}, right). Across the seven models, the spread in giving up halves (standard deviation 6.2 to 3.2 points), and the spread in long loops and in grounded answers falls by about a third (9.7 to 6.4 and 12.9 to 8.1); the spread in reaching a gold source falls only from 17 to 14 points, and in accuracy from 20 to 18. The layer removes the failure modes that separate the weaker models from the strong ones (giving up, looping and answering without grounding) without costing the strong ones much, but it does not close the gap in finding and using the right data, which is where accuracy is decided.

\section{Compaction: Further Results}
\label{app:compaction}

This section backs the compaction results of \Cref{sec:compress}: how it changes the kind of run, where its gains come from, why some agents loop, and what it does to long runs.

\begin{figure*}[tp]
  \centering
  \includegraphics[width=\textwidth]{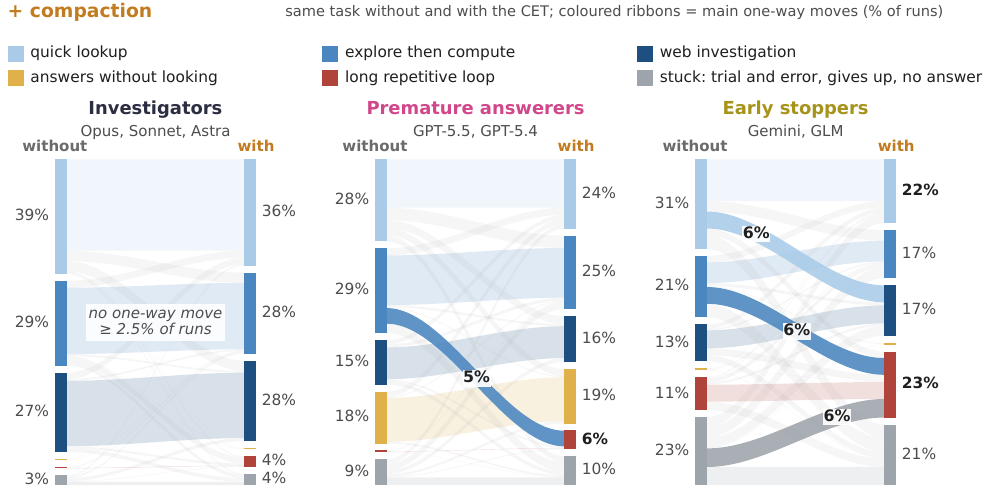}
  \caption{\textbf{Compaction changes little for investigators and pushes the weaker groups into long repetitive runs.} Behavior types of runs without and with compaction, by group.}
  \label{fig:comp-flow}
\end{figure*}

\paragraph{Compaction pushes the weaker groups into loops.} Compaction changes little for the investigators: no move between behavior types reaches 2.5\% of their runs (\Cref{fig:comp-flow}). For the weaker groups it pushes runs into loops: 5\% of the premature answerers' runs and 6\% of the early stoppers' move from exploring then computing into long repetitive runs, so that the early stoppers' long repetitive runs rise from 11\% to 23\%.

\paragraph{Where compaction helps, it helps for different reasons.} Its two clearest gains have different sources. GPT-5.4 gains on DataClaw (+6.4 points, +14.7 for the baseline agent) at every task length, and fewer ``no relevant data'' replies explain only a small part of it (+1.5). GLM gains on HDR mostly where its full transcript had overflowed its context window (+7.1 points there against +0.9 elsewhere, although the difference is not significant), which is the problem compaction is built for. Agents also use what the memory keeps: on DataClaw, when the right value had appeared in a tool output, the premature answerers submit it more often with compaction (wrong answers in those runs fall by 4 points). The cost comes from the rewrites: runs make 1.7 to 3.1 times the calls, and rewritten prompts no longer reuse the provider's cache.

\paragraph{Loops follow the task, not the agent's habits.} The loops have two causes. The premature answerers repeat to re-check a result that is still in their memory (81\% of their repeated actions); the early stoppers repeat to fetch again what a rewrite dropped (only 14\% of their repeats re-check a kept result). Loops grow with the length of the work (for GPT-5.5, 22\% of runs on tasks of one to three steps against 70\% on tasks of eleven or more), and whether a run loops is predicted by how long the baseline run on the same task was and how much computation the task needs (AUC .79 and .76), not by the baseline agent's habits (.50). A task that makes the model hold many intermediate values is what starts the loop.

\paragraph{On very long runs, compaction trades accuracy for tokens.} On runs with very large prompts, compaction trades accuracy for tokens: where the full-transcript prompt exceeds 60k tokens per call, it halves the tokens but lowers accuracy by 3.3 points [$-$5.0, $-$1.6], mostly for Astra, GLM, GPT-5.5 and Gemini. This agrees with studies that find LLM summaries no more accurate than simply dropping old observations \citep{lindenbauer2025complexity,zeng2026loca}.

\section{Combinations and Task Difficulty}
\label{app:combine}

\begin{figure}[tp]
  \centering
  \includegraphics[width=\columnwidth]{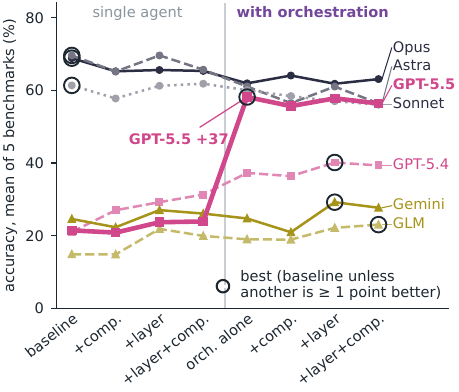}
  \caption{\textbf{The best configuration depends on the model.} Accuracy of the eight configurations (mean of five benchmarks), one line per model; rings mark each model's best configuration (a gain under one point counts as the baseline agent).}
  \label{fig:configs}
\end{figure}

\begin{figure*}[tp]
  \centering
  \includegraphics[width=\textwidth]{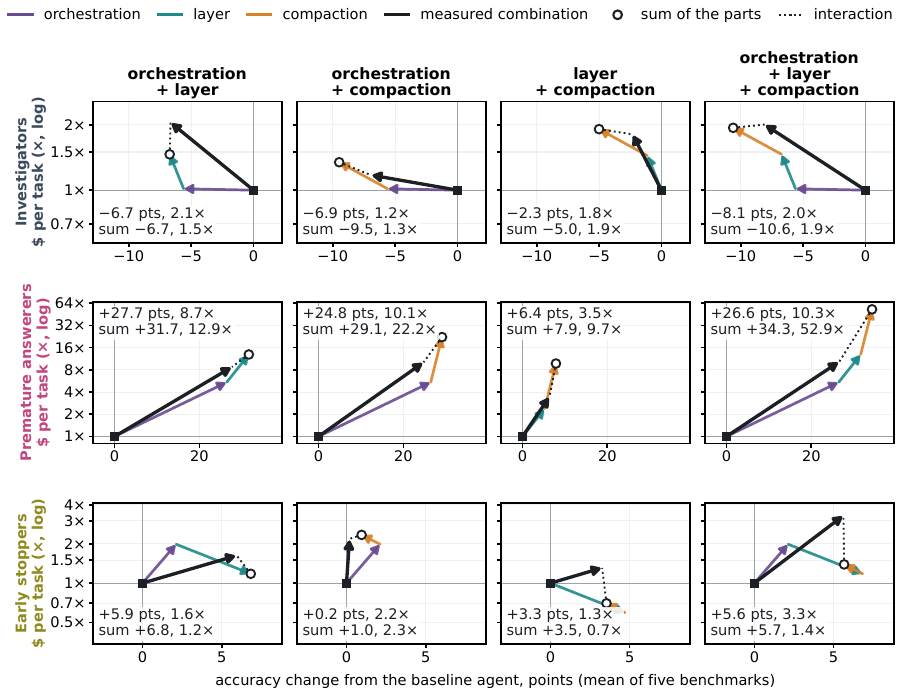}
  \caption{\textbf{CETs that repair the same habit add up to less than their parts.} Each CET as a vector from the baseline agent (square): accuracy change in points (mean of five benchmarks) and dollars per task as a multiple of the baseline agent's (log scale). For each combination (columns) and group (rows), the parts are chained head to tail in color, their sum is the open circle, and the black arrow is the measured combination; the dotted segment is the interaction. Groups are the mean of their models (geometric mean for dollars).}
  \label{fig:vectors}
\end{figure*}

\begin{figure*}[tp]
  \centering
  \includegraphics[width=0.93\textwidth]{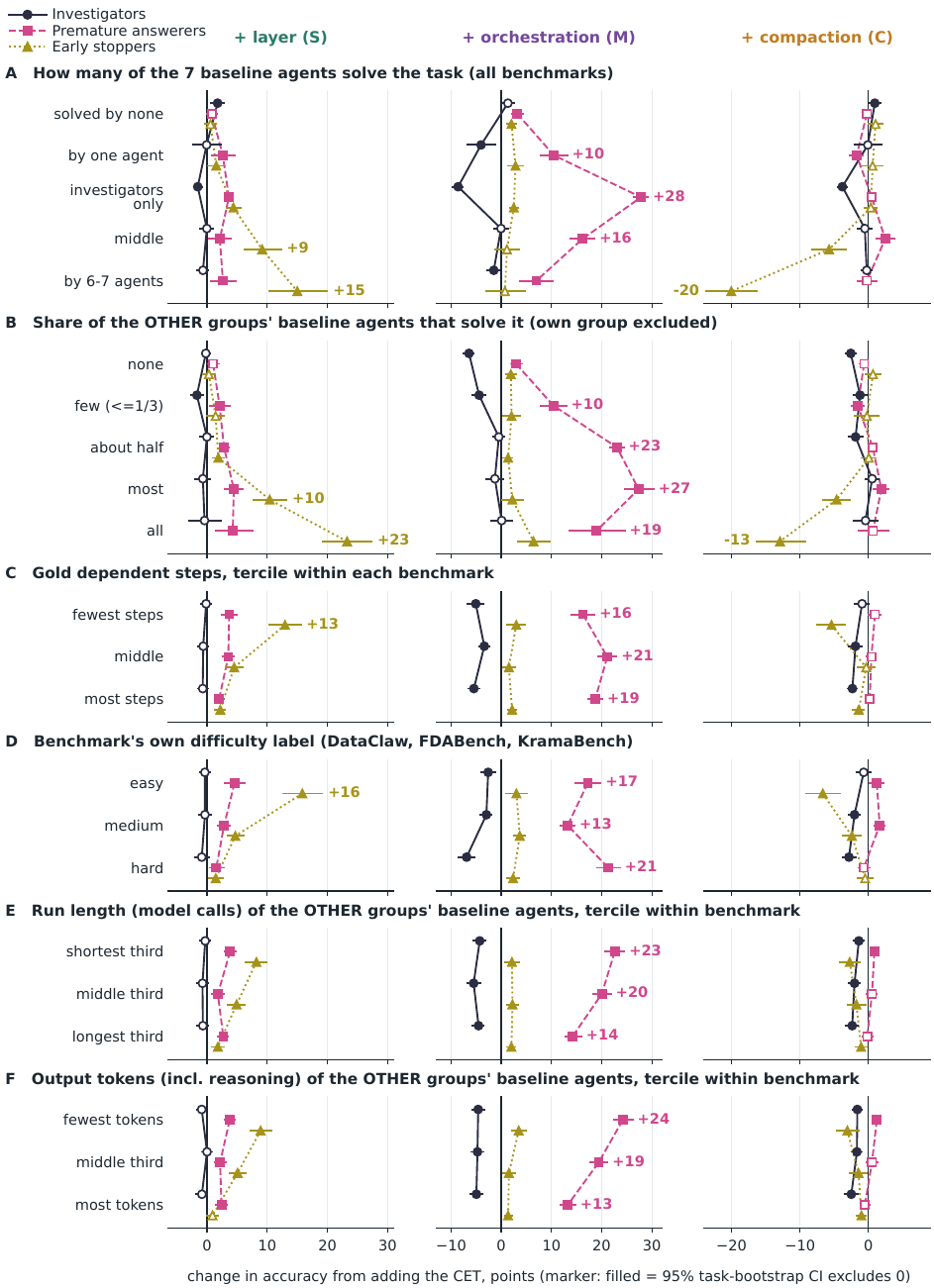}
  \caption{\textbf{The CETs change accuracy where some agent already succeeds, not where the gold says a task is long.} Effect of each CET (averaged over the settings of the other two) by task difficulty, for each group: (A) how many of the seven baseline agents solve the task; (B) the share of the other groups' baseline agents that solve it, which excludes the group's own runs; (C) terciles of the gold number of dependent steps within each benchmark; (D) the benchmarks' own easy, medium and hard labels; (E) terciles of the other groups' baseline run length in model calls and (F) in output tokens, within each benchmark. Filled markers have a 95\% task-bootstrap interval that excludes 0.}
  \label{fig:difficulty-buckets}
\end{figure*}

This section backs \Cref{sec:combine}. It shows the accuracy of every configuration for every model (\Cref{fig:configs}), whether the CETs add up when combined (\Cref{fig:vectors}), and which tasks the CETs change (\Cref{fig:difficulty-buckets}).

\paragraph{The best configuration depends on the model.} \Cref{fig:configs} draws \Cref{tab:cells} as one line per model: the investigators' lines fall from the baseline agent to the right, the premature answerers' lines jump at orchestration, and the early stoppers' lines rise most where the layer is on. The best configurations are orchestration alone for GPT-5.5 (21 to 58), orchestration with the layer for GPT-5.4 (21 to 40) and Gemini (25 to 29), and all three for GLM (15 to 23).

\paragraph{CETs add up only when they repair different habits.} \Cref{fig:vectors} draws each CET as a vector from the baseline agent (accuracy change, dollar ratio), chains the vectors of a combination's parts head to tail, and compares their sum with the measured combination. Two CETs add up only when they repair different habits. For the premature answerers, orchestration and the layer repair the same one, so every combination gains less than its parts (orchestration and layer +27.7 points against a sum of +31.7; all three +26.6 against +34.3): once orchestration makes them look, the layer adds little. The cost behaves differently: all three together cost 10.3 times the baseline agent's dollars, far below the 52.9 the parts would multiply to, partly because with the layer GPT-5.5 stops the re-verification loops it falls into under compaction. Stacking can therefore be cheaper than its parts suggest, even when it is not more accurate. For the early stoppers the opposite holds: accuracy nearly adds up (orchestration and layer +5.9 against +6.8; all three +5.6 against +5.7), but all three cost 3.3 times the baseline agent's dollars against 1.4 for the product of the parts, so most of the combined gain is already bought by the cheap layer alone. For the investigators, combinations lose less than the sum of their losses (all three $-$8.1 against $-$10.6): the CETs damage the same habit, so adding a second one does not double the harm. In practice, the parts' effects are a poor guide to a combination: start from the CET that repairs the agent's main habit, add a second only if it repairs a different one, and measure the combination, since its cost can move either way.

\paragraph{The layer makes compaction safer, and orchestration and compaction combine worst.} Measuring each CET's effect within every setting of the other two, averaged over the seven models, shows where combining helps. Accuracy gains shrink when another CET is on: orchestration adds 5.6 points without the layer and 4.2 with it, and the layer adds 2.7 without orchestration and 1.3 with it; for GPT-5.4 the two shrink from +12.8 and +6.2 alone to +9.5 and +2.9 together. The layer reduces compaction's side effects: compaction raises the share of runs with 100 or more calls by 9.4 points without the layer and 6.2 with it (10.7 against 1.6 for GPT-5.5, 34.6 against 25.4 for GLM), and multiplies calls by 4.1 without the layer and 3.0 with it (9.4 against 1.8 for GPT-5.5). Under orchestration, compaction lowers the share of grounded answers by 6.0 points without the layer and 0.5 with it, in five of the seven models (GPT-5.5 $-$12.6 against +0.1, Gemini $-$10.9 against $-$2.0); Astra is the exception ($-$9.7 against $-$8.8). Orchestration and compaction combine worst: compaction raises the share of runs with 100 or more calls by 10.0 points under orchestration and 5.6 without it (more in six of the seven models), and orchestration raises the share of grounded answers by 12.3 points without compaction and 4.3 with it. With orchestration, the layer cuts calls to 0.85 of their level but raises prompt tokens 1.9 times. The layer's own effects grow with compaction: it cuts repeated actions by 1.9 points alone and 9.6 with orchestration and compaction, and calls to 0.85 and 0.67 of their level without it.

\paragraph{CETs help where some agent already succeeds.} \Cref{fig:difficulty-buckets} sorts tasks by six measures of difficulty: how many baseline agents solve a task (A, B), how long the gold solution is (C), the benchmark's own label (D), and how long other agents' runs on it are (E, F). Three trends hold across them. First, the CETs change accuracy only where some agent already succeeds: on tasks no baseline agent solves, none adds more than 3 points in any group. Second, where they act depends on the group. Orchestration acts on the tasks only the investigators solve as baseline agents (at least two of the three investigators and at most one other model; 52\% of all tasks): these carry 76\% of its gain for the premature answerers and 94\% of its loss for the investigators, and it closes 68\% of the gap between the two groups there, against 12\% for the layer and 14\% for compaction. The layer and compaction act for the early stoppers on the tasks most agents solve, in opposite directions: on tasks six or seven baseline agents solve, the layer adds 15 points and compaction costs 20. Third, length matters much less than solvability: by gold steps or benchmark labels, orchestration's gain for the premature answerers stays at +13 to +21 points at every level, and by the length of other agents' runs the gains of orchestration and the layer shrink on the longest tasks but stay positive. On the most common remaining failure, a wrong value after looking, the CETs repair 8-24\% of cases, about as many as running the same baseline agent again (9-20\%). A CET helps with what some agent can already do; none makes a task solvable that no agent solves.

\begin{figure*}[tp]
  \centering
  \includegraphics[width=\textwidth]{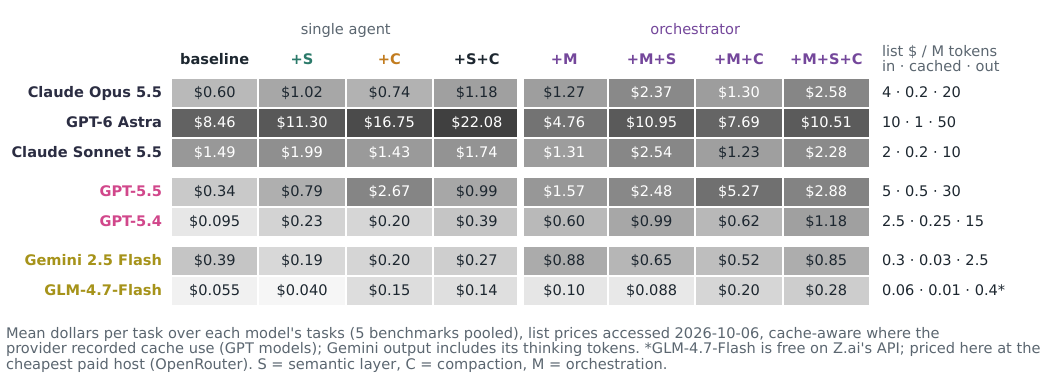}
  \caption{\textbf{Dollars per task by model and configuration.} Pooled over the five benchmarks at list prices (right column: input, cached input and output dollars per million tokens); darker cells cost more.}
  \label{fig:dollars-models}
\end{figure*}

\section{Cost}
\label{app:cost}

This section explains how cost is computed and gives the cost of every model and configuration (\Cref{fig:dollars-models}). Cost is dollars per task at each provider's list price, accessed on 6 October 2026; prices range from \$0.06 to \$10 per million input tokens and from \$0.40 to \$50 per million output tokens, and are printed beside each model in \Cref{fig:dollars-models}. By group, orchestration multiplies dollars per task by 1.7, 4.0 and 2.3 for investigators, premature answerers and early stoppers, the semantic layer by 1.9, 1.4 and 1.1, and compaction by 1.2, 1.4 and 1.8; the baseline agent costs \$1.97, \$0.18 and \$0.15 per task. Input and output tokens are billed at the model's own rates, and reasoning tokens as output. Where the provider recorded prompt-cache use (GPT-5.4, GPT-5.5 and GPT-6 Astra), cached tokens are billed at the cache price, using cache shares measured in a stratified sample of 2,800 runs; Claude recorded no cache use, and the usage records of Gemini and GLM have no cache field, so their cost is an upper bound. Group ratios and baseline costs are geometric means over a group's models, and prices are public list prices. GLM-4.7-Flash, which we served locally and which is free on its provider's API, is priced at the cheapest paid host. Billing all input at list price, or every repeated prompt prefix at the cache price, changes no group's direction except the layer for early stoppers, which becomes cheaper (0.93 times) under the second bound.

\Cref{fig:dollars-models} shows how far the models are apart: a baseline Astra run costs \$8.46 per task and a baseline GLM run \$0.055. Within a model, a CET can change the cost several-fold: compaction raises GPT-5.5 from \$0.34 to \$2.67 through its re-verification loops, and orchestration with compaction to \$5.27, while the layer halves Gemini's baseline cost (\$0.39 to \$0.19).

\section{Benchmark Metrics}
\label{app:published}

This section reports the system-level metrics of each benchmark's paper, computed on our runs. \Cref{tab:native-krama,tab:native-dataclaw,tab:native-fdareport,tab:native-fdachoice,tab:native-hdr} give, for every model, the baseline agent's value of each metric and the change each CET makes, averaged over the settings of the other two, as in the main text. Metrics are scoped as their papers scope them: DataClaw's execution efficiency on correct tasks only, its goal progress and temporal progress efficiency on incorrect tasks only. Where a benchmark releases its evaluator we use it (FDABench's DAG evaluator for the tool orchestration score); where the paper's metric needs an LLM judge (KramaBench pipeline design, DataClaw goal progress and progress efficiency), GPT-5.4 judges with the benchmark's prompt. 

\paragraph{KramaBench (\Cref{tab:native-krama}).} The layer raises the score by 4.3 points on average over the seven models, and most in two domains, legal (+10.1) and wildfire (+7.3), in six of seven models each; for the early stoppers it adds 22.7 points on biomedical tasks. The benchmark's process measure, pipeline design, rises with the layer in six of seven models (+3.2) and with orchestration by 27.7 points for the premature answerers, whose baseline agents implement few of the expected steps (13\%). The layer also cuts the early stoppers' calls by 26 per task.

\begin{table*}[tp]
\centering
\scriptsize
\setlength{\tabcolsep}{3pt}
\resizebox{\textwidth}{!}{%
\begin{tabular}{llrrrrrrrrrrrr}
\toprule
 & & Score & Arch. & Astro. & Biomed. & Env. & Legal & Wildf. & Pipeline & Calls & In (k) & Out (k) & Time (s) \\
\midrule
\multirow{4}{*}{Opus 5.5} & baseline & 82.4 & 50.0 & 75.0 & 83.3 & 83.3 & 90.0 & 93.1 & 35.8 & 3.8 & 47.8 & 1.9 & 104 \\
 & +S & $-$0.9 & +0.0 & $-$2.1 & $-$3.0 & $-$8.8 & +1.7 & +3.8 & +3.7 & $-$1.1 & +73.1 & $-$0.5 & $-$19 \\
 & +M & +0.5 & +0.0 & +6.2 & +8.1 & $-$1.2 & $-$1.7 & $-$1.2 & +5.7 & +2.0 & +39.1 & +4.7 & +42 \\
 & +C & +0.4 & +0.0 & +2.1 & $-$5.4 & +1.2 & +0.0 & +2.0 & +2.1 & +1.2 & $-$13.3 & +1.5 & +31 \\
\cmidrule(lr){1-14}
\multirow{4}{*}{GPT-6 Astra} & baseline & 81.0 & 50.0 & 75.0 & 88.9 & 88.3 & 80.0 & 93.1 & 45.7 & 3.5 & 35.4 & 1.5 & 49 \\
 & +S & $-$1.4 & +0.0 & $-$6.2 & $-$2.2 & $-$4.1 & $-$0.8 & +2.8 & +0.1 & $-$1.5 & +57.4 & $-$0.6 & +9 \\
 & +M & $-$6.0 & +0.0 & $-$16.7 & $-$8.9 & +2.6 & $-$2.5 & $-$15.5 & +0.8 & +2.9 & +67.3 & +2.1 & +63 \\
 & +C & $-$2.2 & +0.0 & $-$4.2 & $-$3.3 & $-$1.7 & $-$0.8 & $-$4.3 & +0.9 & +1.3 & $-$27.1 & +1.6 & +34 \\
\cmidrule(lr){1-14}
\multirow{4}{*}{Sonnet 5.5} & baseline & 78.5 & 50.0 & 75.0 & 81.5 & 88.3 & 76.7 & 88.9 & 36.2 & 5.5 & 87.3 & 3.5 & 213 \\
 & +S & +0.9 & $-$6.2 & $-$1.0 & $-$4.6 & $-$1.8 & +6.7 & +2.6 & $-$1.3 & $-$1.2 & +116 & $-$0.6 & $-$43 \\
 & +M & $-$2.4 & $-$2.1 & $-$11.5 & $-$2.8 & $-$5.8 & +0.0 & +2.3 & +9.0 & $-$0.5 & $-$3.3 & +1.4 & $-$53 \\
 & +C & $-$1.9 & $-$6.2 & $-$5.2 & $-$2.8 & $-$0.8 & +0.0 & $-$1.1 & $-$1.8 & +1.6 & $-$24.4 & +2.6 & +48 \\
\midrule
\multirow{4}{*}{GPT-5.5} & baseline & 22.2 & 4.2 & 41.7 & 3.7 & 12.5 & 20.4 & 40.9 & 2.1 & 1.7 & 12.5 & 1.2 & 147 \\
 & +S & +4.8 & +4.2 & +2.6 & $-$7.7 & $-$1.7 & +12.3 & +7.4 & +2.0 & $-$0.3 & +53.9 & $-$0.3 & $-$343 \\
 & +M & +45.4 & +31.2 & +29.7 & +57.2 & +55.5 & +51.4 & +39.0 & +43.1 & +4.2 & +71.1 & +5.2 & +290 \\
 & +C & +0.1 & +0.0 & $-$2.6 & $-$4.3 & +5.4 & $-$0.4 & $-$0.9 & +0.6 & +0.2 & $-$24.0 & +0.9 & $-$8 \\
\cmidrule(lr){1-14}
\multirow{4}{*}{GPT-5.4} & baseline & 36.5 & 25.0 & 25.0 & 24.6 & 45.8 & 29.8 & 55.5 & 24.9 & 2.6 & 20.2 & 1.4 & 131 \\
 & +S & +7.0 & $-$1.0 & +10.9 & $-$5.2 & $-$4.0 & +14.8 & +13.9 & +4.3 & $-$1.7 & +38.2 & $-$1.7 & $-$625 \\
 & +M & +15.3 & +11.5 & +24.5 & +6.7 & +5.0 & +27.5 & +8.2 & +12.3 & +4.3 & +72.2 & +4.5 & +593 \\
 & +C & +0.8 & $-$3.1 & +3.6 & +3.3 & +2.2 & +1.4 & $-$1.9 & $-$3.1 & +1.3 & $-$13.6 & +1.6 & +130 \\
\midrule
\multirow{4}{*}{Gemini 2.5 Fl.} & baseline & 49.9 & 41.7 & 41.7 & 22.2 & 62.4 & 39.1 & 74.9 & 32.2 & 7.1 & 180 & 13.7 & 180 \\
 & +S & +3.8 & $-$2.1 & $-$8.3 & +23.6 & $-$10.3 & +17.5 & $-$0.5 & +3.9 & $-$4.9 & +247 & $-$7.0 & $-$47 \\
 & +M & $-$6.2 & $-$2.1 & $-$8.3 & +1.4 & $-$19.7 & $-$3.3 & $-$1.9 & $-$3.9 & +12.4 & +585 & +16.0 & +131 \\
 & +C & $-$0.4 & $-$2.1 & $-$2.1 & $-$6.9 & $-$3.2 & +4.2 & +0.7 & +1.1 & +7.4 & $-$94.7 & +11.4 & +118 \\
\cmidrule(lr){1-14}
\multirow{4}{*}{GLM-4.7-Fl.} & baseline & 21.0 & 25.0 & 27.3 & 33.3 & 17.0 & 17.1 & 19.6 & 18.7 & 7.4 & 136 & 16.6 & 7027 \\
 & +S & +15.7 & $-$6.2 & +12.5 & +21.9 & +17.8 & +18.5 & +21.5 & +10.0 & $-$47.8 & +227 & $-$99.7 & $-$10894 \\
 & +M & +3.3 & +14.6 & $-$14.8 & $-$11.5 & +4.1 & +2.1 & +13.6 & $-$2.1 & +20.5 & +595 & $-$10.0 & $-$4842 \\
 & +C & $-$0.4 & +2.1 & +1.1 & $-$5.2 & $-$6.8 & $-$2.9 & +9.2 & $-$1.0 & +89.2 & +1019 & +204 & +16830 \\
\bottomrule
\end{tabular}}
\caption{\textbf{KramaBench in its own metrics.} End-to-end score overall and in each of the six domains, pipeline design (share of the human-annotated key functionalities an LLM judge finds in the executed code, judged by GPT-5.4 with the benchmark's prompt), LLM calls, input and output tokens (thousands) and runtime per task. Each model's first row is the baseline agent; the next three rows give the change made by the layer (+S), orchestration (+M) and compaction (+C), each averaged over the settings of the other two. Rates are in points.}
\label{tab:native-krama}
\end{table*}

\paragraph{DataClaw (\Cref{tab:native-dataclaw}).} The benchmark's process measures show the layer's efficiency more clearly than accuracy does. With the layer, correct runs take fewer steps relative to the gold solution in all seven models (execution efficiency +25 points, +27 even for the investigators, whose accuracy it does not raise), and failed runs reach their milestones earlier (temporal progress efficiency +15, seven of seven); every group makes fewer calls on both correct and wrong tasks. Orchestration lets even failed runs get further: goal progress on incorrect tasks rises in all seven models (+6.5 points; investigators +9.6). The early stoppers' layer gain sits on easy tasks (+26.7 points) and vanishes on hard ones ($-$1.0). Compaction lowers execution efficiency in all seven models ($-$8.6), partly because the metric counts each memory rewrite as a step.

\begin{table*}[tp]
\centering
\scriptsize
\setlength{\tabcolsep}{3pt}
\resizebox{\textwidth}{!}{%
\begin{tabular}{llrrrrrrrrrrr}
\toprule
 & & Acc & Easy & Med. & Hard & EE & GPR & TPE & Calls (C) & Calls (W) & In (k) & Out (k) \\
\midrule
\multirow{4}{*}{Opus 5.5} & baseline & 70.8 & 91.6 & 67.8 & 46.0 & 57.9 & 40.9 & 88.7 & 6.8 & 8.0 & 68.3 & 3.1 \\
 & +S & $-$2.3 & +0.2 & $-$2.1 & $-$7.1 & +29.7 & $-$5.5 & +9.5 & $-$5.3 & $-$6.2 & +381 & $-$4.3 \\
 & +M & +1.5 & +0.6 & +1.8 & +2.0 & $-$16.0 & +12.1 & $-$7.7 & +7.9 & +11.2 & +321 & +12.7 \\
 & +C & $-$0.3 & +0.2 & $-$0.8 & +0.5 & $-$3.8 & $-$0.2 & +0.0 & +6.6 & +8.5 & +44.9 & +12.1 \\
\cmidrule(lr){1-13}
\multirow{4}{*}{GPT-6 Astra} & baseline & 71.9 & 90.8 & 68.5 & 52.0 & 65.7 & 44.3 & 93.7 & 5.8 & 7.2 & 79.2 & 1.6 \\
 & +S & $-$1.2 & $-$0.2 & $-$0.6 & $-$5.5 & +27.0 & +0.1 & +7.0 & $-$9.5 & $-$35.2 & +222 & $-$35.2 \\
 & +M & $-$2.0 & $-$2.5 & $-$0.8 & $-$5.3 & $-$10.4 & +7.7 & $-$9.1 & +11.1 & +44.7 & +416 & +44.7 \\
 & +C & $-$0.1 & $-$2.1 & +1.3 & $-$2.1 & $-$13.1 & +1.7 & $-$12.1 & +15.5 & +56.1 & +300 & +67.9 \\
\cmidrule(lr){1-13}
\multirow{4}{*}{Sonnet 5.5} & baseline & 65.7 & 92.2 & 61.0 & 37.2 & 28.1 & 40.5 & 39.3 & 28.4 & 38.6 & 436 & 23.4 \\
 & +S & $-$1.3 & $-$2.5 & $-$0.2 & $-$3.4 & +25.4 & +1.6 & +35.7 & $-$25.3 & $-$33.0 & +907 & $-$29.9 \\
 & +M & +1.3 & +0.2 & +1.8 & +1.7 & $-$2.1 & +9.0 & $-$3.0 & +10.3 & +20.5 & +651 & +20.6 \\
 & +C & $-$2.3 & $-$1.4 & $-$2.9 & $-$1.9 & $-$3.2 & $-$6.4 & +1.7 & +17.2 & +21.1 & $-$123 & +35.8 \\
\midrule
\multirow{4}{*}{GPT-5.5} & baseline & 62.9 & 90.9 & 56.3 & 39.1 & 58.1 & 42.3 & 89.5 & 6.6 & 8.0 & 109 & 4.1 \\
 & +S & +0.9 & $-$1.0 & +1.2 & +3.0 & +44.8 & +0.5 & +15.7 & $-$60.3 & $-$108.5 & $-$6.3 & $-$91.5 \\
 & +M & +0.4 & +0.0 & +1.1 & $-$1.4 & +4.0 & +0.1 & $-$1.5 & +15.0 & +31.0 & +233 & +30.6 \\
 & +C & $-$2.7 & $-$2.1 & $-$1.8 & $-$7.5 & $-$23.9 & $-$0.7 & $-$15.0 & +65.2 & +121.8 & +354 & +113 \\
\cmidrule(lr){1-13}
\multirow{4}{*}{GPT-5.4} & baseline & 28.2 & 52.4 & 22.4 & 7.9 & 59.2 & 14.3 & 96.0 & 5.8 & 5.8 & 50.4 & 2.1 \\
 & +S & +6.7 & +10.8 & +6.1 & +1.9 & +22.7 & +5.6 & +5.0 & $-$2.7 & $-$1.7 & +295 & $-$0.2 \\
 & +M & +6.7 & +10.9 & +6.6 & $-$0.4 & $-$1.1 & +7.5 & $-$13.2 & +6.2 & +14.7 & +277 & +11.7 \\
 & +C & +6.4 & +7.0 & +7.2 & +2.0 & $-$1.7 & +3.5 & $-$1.6 & +5.3 & +11.3 & +85.2 & +14.2 \\
\midrule
\multirow{4}{*}{Gemini 2.5 Fl.} & baseline & 22.2 & 42.9 & 17.8 & 3.0 & 37.7 & 7.9 & 64.4 & 11.6 & 14.7 & 2986 & 12.2 \\
 & +S & +13.4 & +24.5 & +11.7 & +0.6 & +12.3 & +8.5 & +16.7 & $-$21.5 & $-$27.7 & $-$514 & $-$30.6 \\
 & +M & +2.7 & +3.2 & +2.3 & +3.2 & $-$16.7 & +7.4 & $-$19.3 & +43.8 & +58.5 & +1817 & +61.0 \\
 & +C & $-$5.3 & $-$9.7 & $-$4.0 & $-$2.2 & $-$6.8 & $-$3.4 & $-$9.2 & +41.2 & +58.4 & $-$1629 & +64.7 \\
\cmidrule(lr){1-13}
\multirow{4}{*}{GLM-4.7-Fl.} & baseline & 15.1 & 30.6 & 10.1 & 6.8 & 20.5 & 15.8 & 38.7 & 37.2 & 49.4 & 1947 & 45.8 \\
 & +S & +10.6 & +28.8 & +5.6 & $-$2.6 & +16.1 & +10.1 & +14.1 & $-$68.9 & $-$76.5 & +799 & $-$121 \\
 & +M & +4.2 & +4.9 & +5.0 & $-$0.2 & $-$0.8 & +1.6 & $-$6.1 & +31.8 & +37.6 & +1257 & +91.6 \\
 & +C & $-$5.8 & $-$13.4 & $-$3.3 & $-$2.4 & $-$7.6 & $-$6.1 & $-$13.6 & +99.7 & +156.7 & +450 & +367 \\
\bottomrule
\end{tabular}}
\caption{\textbf{DataClaw in its own metrics.} Judged accuracy overall and by difficulty; execution efficiency (EE, gold steps over agent steps, correct tasks only); goal progress rate (GPR, share of gold milestones reached) and temporal progress efficiency (TPE, $\gamma=0.9$), both on incorrect tasks only, judged by GPT-5.4; LLM calls on correct (C) and wrong (W) tasks; input and output tokens (thousands) per task. Each model's first row is the baseline agent; the next three rows give the change made by the layer (+S), orchestration (+M) and compaction (+C), each averaged over the settings of the other two. Rates are in points.}
\label{tab:native-dataclaw}
\end{table*}

\paragraph{FDABench reports (\Cref{tab:native-fdareport}).} Orchestration raises the rubric score by 33.3 points for the premature answerers and lowers it by 5.1 for the investigators. Of the four rubric keys, the layer raises the one that checks correctness, SQL accuracy, in six of seven models (+1.5), and lowers external integration in all seven ($-$3.1): with a description of the database, reports draw more on the database and less on outside evidence. The tool measures show the gap between process and accuracy most sharply: with orchestration, the tool orchestration score rises by 54 points, tool recall by 30 and tool F1 by 24, in all seven models, including the investigators (+53.5 on the orchestration score) whose rubric score falls. ROUGE barely moves, except for the premature answerers' longer orchestrated reports (+11.7 ROUGE-1).

\begin{table*}[tp]
\centering
\scriptsize
\setlength{\tabcolsep}{3pt}
\resizebox{\textwidth}{!}{%
\begin{tabular}{llrrrrrrrrrrrrrrr}
\toprule
 & & RS & SQL & Ext. & Logic & Compl. & R-1 & R-L & TR & TF1 & TOS & SR & Calls & In (k) & Out (k) & Time (s) \\
\midrule
\multirow{4}{*}{Opus 5.5} & baseline & 59.8 & 32.7 & 76.3 & 85.7 & 45.6 & 36.5 & 14.2 & 57.3 & 70.6 & 35.5 & 97.9 & 4.7 & 102 & 4.0 & 133 \\
 & +S & $-$1.3 & $-$0.1 & $-$6.4 & $-$1.6 & $-$0.2 & +0.3 & +0.1 & $-$4.6 & $-$3.7 & $-$2.5 & +0.4 & $-$2.8 & +90.8 & $-$3.9 & +44 \\
 & +M & $-$8.1 & $-$4.9 & $-$8.0 & $-$4.7 & $-$7.9 & $-$1.4 & $-$0.7 & +25.6 & +19.7 & +55.3 & $-$1.2 & +13.9 & +210 & +28.6 & +364 \\
 & +C & +1.5 & $-$2.5 & $-$0.2 & +0.3 & +1.6 & $-$0.4 & $-$0.2 & +0.1 & +0.2 & +0.3 & $-$1.4 & +4.9 & $-$53.5 & +9.6 & +135 \\
\cmidrule(lr){1-17}
\multirow{4}{*}{GPT-6 Astra} & baseline & 59.4 & 25.2 & 83.7 & 86.7 & 42.0 & 33.4 & 12.6 & 72.0 & 82.6 & 36.4 & 96.2 & 9.7 & 241 & 7.7 & 545 \\
 & +S & +0.8 & +3.6 & $-$2.4 & +0.5 & +1.7 & +0.4 & +0.1 & +2.0 & +1.4 & +1.5 & $-$1.0 & +2.3 & +425 & +3.0 & +1605 \\
 & +M & $-$3.9 & $-$2.4 & $-$8.2 & $-$3.6 & $-$1.4 & +0.1 & $-$0.1 & +12.1 & +7.5 & +52.8 & $-$1.6 & $-$17.0 & $-$153 & $-$64.4 & +172 \\
 & +C & $-$8.1 & $-$0.6 & $-$15.4 & $-$9.5 & $-$6.9 & $-$0.2 & +0.2 & +1.0 & +0.6 & $-$1.1 & $-$0.9 & +44.6 & +253 & +97.2 & +3945 \\
\cmidrule(lr){1-17}
\multirow{4}{*}{Sonnet 5.5} & baseline & 46.3 & 25.2 & 54.6 & 74.1 & 31.2 & 34.8 & 13.9 & 58.5 & 72.2 & 36.0 & 97.1 & 5.5 & 126 & 4.8 & 184 \\
 & +S & $-$0.7 & +0.6 & $-$3.4 & +0.8 & +1.0 & +0.1 & +0.0 & $-$0.9 & $-$0.8 & $-$0.3 & $-$0.4 & $-$1.6 & +195 & $-$1.1 & +17 \\
 & +M & $-$3.3 & $-$1.9 & +2.3 & $-$6.7 & $-$3.3 & $-$0.8 & $-$0.3 & +20.3 & +14.6 & +52.2 & $-$0.1 & +10.7 & +222 & +24.2 & +317 \\
 & +C & +0.9 & +0.5 & $-$3.3 & +2.7 & +1.4 & $-$0.3 & $-$0.2 & $-$0.4 & +0.1 & +0.7 & $-$1.8 & +5.6 & $-$51.2 & +12.6 & +158 \\
\midrule
\multirow{4}{*}{GPT-5.5} & baseline & 9.9 & 12.4 & 1.7 & 21.4 & 3.9 & 17.4 & 7.6 & 26.3 & 37.6 & 24.5 & 59.5 & 11.2 & 150 & 4.9 & 1716 \\
 & +S & +0.4 & +0.0 & $-$0.5 & +0.7 & +1.4 & $-$0.3 & $-$0.1 & $-$2.3 & $-$2.1 & $-$0.9 & $-$2.1 & +1.4 & +172 & +1.6 & +377 \\
 & +M & +46.6 & +13.4 & +71.9 & +63.0 & +38.2 & +20.8 & +7.2 & +57.4 & +53.5 & +65.5 & +24.0 & +3.4 & +89.8 & +19.7 & +444 \\
 & +C & $-$0.7 & $-$0.9 & $-$1.7 & +0.2 & $-$0.3 & $-$0.8 & $-$0.3 & $-$0.7 & $-$0.6 & $-$0.7 & $-$2.0 & +8.8 & $-$58.7 & +14.2 & +110 \\
\cmidrule(lr){1-17}
\multirow{4}{*}{GPT-5.4} & baseline & 27.9 & 10.1 & 25.7 & 59.6 & 16.3 & 33.6 & 12.9 & 27.4 & 39.3 & 22.8 & 59.8 & 2.6 & 31.7 & 2.0 & 2145 \\
 & +S & +2.4 & +5.1 & $-$0.5 & +1.8 & +3.4 & +0.4 & +0.2 & +3.8 & +5.6 & +5.0 & $-$2.6 & +0.3 & +93.8 & +0.3 & +708 \\
 & +M & +19.9 & +8.5 & +33.9 & +18.1 & +19.2 & +2.7 & +0.8 & +50.4 & +44.1 & +61.5 & +20.8 & +18.1 & +225 & +19.9 & $-$1169 \\
 & +C & +0.4 & +2.8 & $-$3.9 & +2.4 & +0.3 & +0.0 & $-$0.1 & +0.8 & +0.9 & +0.1 & +2.7 & +4.9 & $-$65.5 & +4.8 & +72 \\
\midrule
\multirow{4}{*}{Gemini 2.5 Fl.} & baseline & 24.5 & 10.1 & 28.4 & 48.8 & 15.4 & 36.8 & 15.4 & 51.0 & 65.9 & 38.5 & 67.3 & 10.9 & 454 & 12.5 & 218 \\
 & +S & +0.3 & +1.1 & $-$2.1 & $-$0.3 & +0.9 & $-$0.0 & +0.1 & +8.0 & +7.3 & +6.9 & +1.0 & $-$6.0 & +330 & $-$15.7 & +1736 \\
 & +M & +5.4 & $-$4.1 & +19.3 & +5.9 & +5.4 & +3.3 & +1.0 & +20.3 & +14.1 & +42.4 & +16.7 & +60.2 & +1504 & +64.6 & +2712 \\
 & +C & $-$0.4 & $-$0.6 & +0.5 & $-$1.5 & $-$3.0 & $-$0.1 & $-$0.1 & +0.5 & +0.4 & $-$0.1 & +0.2 & +33.9 & $-$925 & +18.8 & +1061 \\
\cmidrule(lr){1-17}
\multirow{4}{*}{GLM-4.7-Fl.} & baseline & 29.2 & 5.0 & 49.8 & 46.1 & 15.8 & 33.1 & 13.0 & 43.7 & 58.6 & 32.7 & 46.0 & 6.1 & 113 & 13.3 & 5199 \\
 & +S & $-$2.0 & +0.1 & $-$6.2 & +1.3 & $-$1.4 & +0.2 & +0.1 & $-$0.1 & +0.5 & +1.2 & $-$4.7 & +1.9 & +486 & +14.7 & $-$2025 \\
 & +M & +4.0 & +6.8 & $-$2.7 & +4.0 & +10.1 & $-$5.4 & $-$2.6 & +20.6 & +15.5 & +46.3 & +9.7 & +64.4 & +617 & +122 & +1341 \\
 & +C & +0.8 & +1.8 & +0.6 & +0.5 & +1.2 & $-$0.3 & $-$0.1 & +10.5 & +8.5 & +4.0 & +1.5 & +70.8 & +504 & +146 & +8472 \\
\bottomrule
\end{tabular}}
\caption{\textbf{FDABench reports in their own metrics.} Rubric score (RS) and its four keys (SQL accuracy, external integration, logical reasoning, completeness), ROUGE-1 and ROUGE-L, tool recall (TR) and tool F1 (TF1) at the level of tool names, tool orchestration score (TOS, the released DAG evaluator), tool success rate (SR, runs that attempted a tool), external LLM calls, input and output tokens (thousands) and latency per task; 218 gold-labelled report tasks. Each model's first row is the baseline agent; the next three rows give the change made by the layer (+S), orchestration (+M) and compaction (+C), each averaged over the settings of the other two. Rates are in points.}
\label{tab:native-fdareport}
\end{table*}

\paragraph{FDABench multiple choice (\Cref{tab:native-fdachoice}).} Orchestration raises exact match by 21.9 points for the premature answerers and lowers it by 11.2 for the investigators, with the same sign at every difficulty. The option measures show how: under orchestration the investigators choose fewer options, losing 6.9 points of recall and 6.7 of precision, while the other two groups choose more of the right ones (recall +14.3 and +12.8). The tool orchestration score rises in all seven models (+30).

\begin{table*}[tp]
\centering
\scriptsize
\setlength{\tabcolsep}{3pt}
\resizebox{\textwidth}{!}{%
\begin{tabular}{llrrrrrrrrrrrrr}
\toprule
 & & EX & Easy & Med. & Hard & Prec. & Rec. & TF1 & TOS & SR & Calls & In (k) & Out (k) & Time (s) \\
\midrule
\multirow{4}{*}{Opus 5.5} & baseline & 70.0 & 65.2 & 66.7 & 73.8 & 93.3 & 88.3 & 75.7 & 32.2 & 99.7 & 6.3 & 129 & 2.6 & 150 \\
 & +S & +0.0 & +1.1 & $-$0.2 & $-$0.2 & $-$0.2 & +0.5 & $-$0.5 & $-$1.1 & +0.1 & $-$0.4 & +100 & +0.2 & +164 \\
 & +M & $-$10.5 & $-$9.8 & $-$9.9 & $-$11.1 & $-$2.5 & $-$6.5 & +15.6 & +22.4 & $-$0.4 & +8.0 & +47.4 & +11.7 & +216 \\
 & +C & $-$3.1 & +4.3 & $-$5.0 & $-$3.9 & $-$1.4 & $-$1.6 & +0.5 & +0.1 & $-$0.4 & +3.2 & $-$41.9 & +6.3 & +90 \\
\cmidrule(lr){1-15}
\multirow{4}{*}{GPT-6 Astra} & baseline & 66.6 & 69.6 & 64.0 & 67.5 & 90.0 & 86.7 & 78.9 & 33.5 & 98.7 & 9.9 & 280 & 2.7 & 601 \\
 & +S & +2.3 & $-$0.5 & +3.7 & +2.0 & $-$0.2 & +1.3 & +0.1 & $-$1.5 & +0.1 & +4.3 & +290 & +5.7 & +1762 \\
 & +M & $-$17.3 & $-$12.5 & $-$16.4 & $-$19.2 & +0.2 & $-$8.8 & +13.6 & +18.8 & $-$0.4 & $-$19.9 & $-$331 & $-$45.9 & +179 \\
 & +C & $-$4.3 & $-$2.7 & $-$4.6 & $-$4.5 & $-$0.4 & $-$3.2 & $-$0.1 & +0.2 & +0.1 & +39.3 & +236 & +65.8 & +3526 \\
\cmidrule(lr){1-15}
\multirow{4}{*}{Sonnet 5.5} & baseline & 55.6 & 43.5 & 54.4 & 60.0 & 90.5 & 83.0 & 76.4 & 32.6 & 99.0 & 7.5 & 158 & 5.2 & 247 \\
 & +S & +0.8 & +4.3 & +0.9 & $-$0.3 & $-$1.3 & $-$0.3 & $-$0.3 & $-$0.2 & +0.0 & +0.6 & +149 & +0.9 & +118 \\
 & +M & $-$5.9 & $-$5.4 & $-$7.0 & $-$5.3 & $-$17.7 & $-$5.5 & +13.8 & +21.5 & $-$0.2 & +5.0 & $-$15.0 & +6.2 & +174 \\
 & +C & $-$5.0 & +1.1 & $-$7.0 & $-$5.3 & $-$2.0 & $-$2.1 & +0.6 & +0.2 & $-$0.3 & +5.1 & $-$41.5 & +8.0 & +166 \\
\midrule
\multirow{4}{*}{GPT-5.5} & baseline & 3.0 & 3.3 & 1.8 & 3.8 & 63.7 & 40.9 & 0.3 & 0.1 & 100.0 & 1.0 & 6.6 & 0.3 & 590 \\
 & +S & $-$0.3 & $-$0.8 & +1.8 & $-$1.6 & +0.7 & +0.3 & +0.1 & $-$0.4 & $-$13.5 & +0.8 & +64.3 & +1.6 & +990 \\
 & +M & +37.4 & +34.5 & +32.0 & +42.0 & +17.1 & +26.7 & +90.6 & +53.7 & +8.9 & +18.1 & +179 & +18.8 & +2756 \\
 & +C & +0.0 & $-$1.4 & +0.9 & $-$0.2 & +1.0 & +1.1 & +0.2 & $-$0.0 & $-$1.0 & +2.8 & $-$44.9 & +2.6 & +321 \\
\cmidrule(lr){1-15}
\multirow{4}{*}{GPT-5.4} & baseline & 4.4 & 2.2 & 3.5 & 5.6 & 63.6 & 51.8 & 25.1 & 8.8 & 66.4 & 1.7 & 11.0 & 0.2 & 356 \\
 & +S & +2.2 & +3.0 & +1.8 & +2.3 & +2.0 & +1.9 & +3.0 & +3.0 & +1.5 & +0.6 & +56.3 & +0.7 & +1416 \\
 & +M & +6.4 & +10.6 & +3.9 & +7.0 & +12.8 & +1.8 & +59.7 & +40.3 & +18.5 & +13.4 & +114 & +9.5 & +646 \\
 & +C & $-$0.5 & +0.3 & $-$1.3 & $-$0.2 & +0.2 & $-$0.5 & +1.4 & +1.0 & +3.2 & +2.8 & $-$15.2 & +1.8 & $-$74 \\
\midrule
\multirow{4}{*}{Gemini 2.5 Fl.} & baseline & 7.5 & 4.3 & 5.3 & 10.0 & 64.2 & 54.7 & 64.7 & 26.6 & 85.7 & 6.9 & 189 & 4.4 & 186 \\
 & +S & +0.1 & +3.3 & +0.4 & $-$1.1 & +2.3 & $-$1.8 & +1.5 & +3.6 & +2.3 & +5.0 & +196 & +4.4 & +2338 \\
 & +M & +4.8 & +5.4 & +4.8 & +4.5 & +1.2 & +3.7 & +20.8 & +30.4 & +5.0 & +18.6 & +114 & +9.0 & +2538 \\
 & +C & $-$1.8 & $-$3.3 & +0.0 & $-$2.7 & $-$3.3 & $-$1.8 & $-$0.1 & $-$0.8 & $-$2.1 & +10.2 & $-$121 & +7.5 & +834 \\
\cmidrule(lr){1-15}
\multirow{4}{*}{GLM-4.7-Fl.} & baseline & 3.4 & 6.7 & 3.5 & 2.5 & 58.5 & 41.2 & 57.5 & 22.6 & 54.3 & 5.9 & 100 & 14.9 & 6126 \\
 & +S & $-$0.2 & $-$2.2 & $-$0.2 & +0.3 & +0.2 & $-$0.1 & +1.0 & +4.3 & $-$0.2 & +2.8 & +161 & +9.1 & +1172 \\
 & +M & +1.0 & +2.2 & +0.2 & +1.2 & $-$10.9 & +21.8 & +14.3 & +24.2 & +9.1 & +41.1 & +230 & +80.0 & +2305 \\
 & +C & +0.1 & $-$1.1 & $-$1.1 & +1.2 & +0.1 & +2.1 & +6.3 & +3.7 & +2.5 & +43.3 & +176 & +94.2 & +1387 \\
\bottomrule
\end{tabular}}
\caption{\textbf{FDABench multiple choice in its own metrics.} Exact match of the option set (EX) overall and by difficulty, precision and recall of the chosen options, tool F1, tool orchestration score (the released evaluator's fallback, as choice tasks have no released DAG), tool success rate (runs that attempted a tool), LLM calls, input and output tokens (thousands) and latency per task; 320 multiple-choice tasks. Each model's first row is the baseline agent; the next three rows give the change made by the layer (+S), orchestration (+M) and compaction (+C), each averaged over the settings of the other two. Rates are in points.}
\label{tab:native-fdachoice}
\end{table*}

\paragraph{HDR (\Cref{tab:native-hdr}).} Orchestration's gain for the premature answerers is largest on parallel tasks (+36.6 points), against +21.6 when SQL comes first and +13.7 when search comes first. With the layer, more of the early stoppers' attempted tool calls succeed (+6.6 points). Under the benchmark's own limits, compaction raises the share of runs that time out in all seven models (+9.7 points; early stoppers +26.2), from 2.2\% for the baseline agent.

\begin{table*}[tp]
\centering
\scriptsize
\setlength{\tabcolsep}{3pt}
\resizebox{\textwidth}{!}{%
\begin{tabular}{llrrrrrrrrrr}
\toprule
 & & Succ. & S2SQL & SQL2S & Par. & LLM calls & Tool calls & Tool SR & In (k) & Out (k) & Timeout \\
\midrule
\multirow{4}{*}{Opus 5.5} & baseline & 61.1 & 50.0 & 54.7 & 77.3 & 7.0 & 5.8 & 99.4 & 264 & 4.9 & 0.0 \\
 & +S & $-$0.8 & $-$0.4 & $-$2.1 & +1.3 & $-$1.6 & $-$1.6 & +0.4 & +197 & $-$1.0 & +0.2 \\
 & +M & $-$1.1 & +5.6 & $-$2.6 & $-$1.7 & +15.1 & +9.4 & $-$0.5 & +94.6 & +18.1 & +0.2 \\
 & +C & +0.9 & +3.9 & +0.1 & +0.8 & +6.5 & +0.2 & $-$0.8 & $-$22.6 & +6.5 & +0.3 \\
\cmidrule(lr){1-12}
\multirow{4}{*}{GPT-6 Astra} & baseline & 68.9 & 63.8 & 61.1 & 84.9 & 32.7 & 32.6 & 99.7 & 2127 & 13.1 & 4.0 \\
 & +S & +0.1 & $-$1.3 & $-$0.6 & +1.9 & $-$1.8 & $-$0.5 & $-$0.0 & +767 & $-$3.4 & $-$0.2 \\
 & +M & $-$14.7 & $-$11.6 & $-$14.4 & $-$16.6 & $-$37.1 & $-$28.4 & $-$0.0 & $-$1743 & $-$87.1 & $-$13.6 \\
 & +C & $-$7.2 & $-$3.0 & $-$9.0 & $-$6.1 & +47.2 & +16.1 & $-$0.1 & $-$219 & +108 & +10.7 \\
\cmidrule(lr){1-12}
\multirow{4}{*}{Sonnet 5.5} & baseline & 60.5 & 53.4 & 54.2 & 74.8 & 21.9 & 14.8 & 99.9 & 1738 & 48.1 & 4.2 \\
 & +S & $-$1.4 & +0.4 & $-$1.1 & $-$2.9 & $-$12.5 & $-$7.4 & $-$0.0 & +385 & $-$20.2 & $-$2.0 \\
 & +M & $-$3.0 & $-$3.0 & $-$3.3 & $-$2.5 & +6.6 & +11.1 & +0.0 & $-$720 & $-$20.8 & $-$1.6 \\
 & +C & +1.3 & $-$1.3 & +2.8 & $-$0.0 & +13.6 & +1.6 & $-$0.3 & $-$526 & +25.3 & +0.9 \\
\midrule
\multirow{4}{*}{GPT-5.5} & baseline & 9.3 & 27.0 & 2.3 & 12.6 & 5.3 & 0.1 & 83.3 & 90.4 & 1.5 & 0.1 \\
 & +S & +1.2 & +1.2 & +0.3 & +2.9 & $-$0.2 & $-$0.4 & +3.9 & +331 & $-$1.5 & +3.5 \\
 & +M & +42.8 & +20.2 & +40.1 & +58.3 & +32.7 & +22.7 & +5.6 & +397 & +38.8 & +8.7 \\
 & +C & $-$2.1 & $-$4.8 & $-$1.6 & $-$1.6 & +9.2 & +2.9 & +1.7 & $-$61.9 & +8.9 & +3.2 \\
\cmidrule(lr){1-12}
\multirow{4}{*}{GPT-5.4} & baseline & 8.1 & 9.2 & 5.3 & 12.3 & 2.4 & 1.4 & 86.6 & 37.1 & 0.4 & 0.4 \\
 & +S & +4.5 & +2.6 & +1.0 & +11.4 & $-$0.5 & $-$0.0 & $-$0.6 & +122 & $-$0.6 & $-$0.4 \\
 & +M & +7.4 & +7.2 & +3.0 & +14.9 & +17.8 & +10.0 & +5.9 & +162 & +12.8 & +3.0 \\
 & +C & +0.9 & $-$2.0 & +0.9 & +2.5 & +5.3 & +2.7 & $-$1.4 & +7.0 & +4.7 & +0.3 \\
\midrule
\multirow{4}{*}{Gemini 2.5 Fl.} & baseline & 18.7 & 6.9 & 9.9 & 39.5 & 9.1 & 7.3 & 80.3 & 241 & 3.8 & 0.0 \\
 & +S & +4.2 & +0.9 & +1.5 & +10.5 & +5.0 & +1.6 & +10.3 & +611 & +3.8 & +10.7 \\
 & +M & $-$3.4 & +2.6 & +0.5 & $-$13.0 & +35.3 & +18.6 & $-$7.3 & +395 & +7.0 & +11.1 \\
 & +C & $-$2.9 & +0.0 & $-$2.0 & $-$5.9 & +28.9 & +10.6 & $-$2.1 & +178 & +15.7 & +7.2 \\
\cmidrule(lr){1-12}
\multirow{4}{*}{GLM-4.7-Fl.} & baseline & 5.5 & 3.4 & 2.0 & 12.7 & 8.0 & 11.4 & 51.7 & 236 & 27.1 & 6.6 \\
 & +S & +0.1 & $-$0.9 & +1.1 & $-$1.1 & $-$9.8 & $-$6.5 & +3.0 & +491 & $-$19.9 & +6.1 \\
 & +M & +2.0 & $-$2.6 & +1.6 & +4.9 & +42.6 & +24.6 & +16.5 & +202 & +74.4 & +28.6 \\
 & +C & +3.8 & $-$0.9 & +1.8 & +9.5 & +73.2 & +43.5 & +1.9 & +727 & +205 & +45.1 \\
\bottomrule
\end{tabular}}
\caption{\textbf{HDR in its own metrics.} Success rate overall and by task type (search then SQL, SQL then search, parallel), LLM calls, tool calls and their success rate (runs that attempted a tool), input and output tokens (thousands) per task, and the share of runs that exceed the benchmark's limits of 300 tool-calling turns or 3,600 seconds. Each model's first row is the baseline agent; the next three rows give the change made by the layer (+S), orchestration (+M) and compaction (+C), each averaged over the settings of the other two. Rates are in points.}
\label{tab:native-hdr}
\end{table*}

\FloatBarrier \section{Implementation and Compute}
\label{app:compute}

\paragraph{Software.} The harness is written in Python 3.10 and calls every API model through an OpenAI-compatible chat-completion interface. Tool code runs in a separate Python process with a time limit, SQL runs on read-only copies of the benchmark databases (SQLite, and BigQuery for the FDABench tasks hosted there), and web search uses Perplexity Sonar\footnote{\url{https://openrouter.ai/perplexity/sonar}}. The semantic layer's dense channel embeds entries with text-embedding-3-small\footnote{\url{https://developers.openai.com/api/docs/models/text-embedding-3-small}}, and its lexical channel scores the term overlap of identifiers, codes and units. The main libraries are vLLM 0.22.1, PyTorch 2.11, pandas 2.3 and NumPy 2.2.

\paragraph{Hardware.} All models except GLM-4.7-Flash are accessed through their providers' APIs, which do not disclose their parameter counts (GPT-5.4, GPT-5.5, GPT-6 Astra, Claude Sonnet 5.5, Claude Opus 5.5, Gemini 2.5 Flash and text-embedding-3-small), so these agents run on CPU nodes (48 cores and 256 GB of memory per job), each job running many tasks in parallel. GLM-4.7-Flash (30B parameters, 3B active) is served with vLLM in BF16 on NVIDIA A100 80GB GPUs, with tensor parallelism of 2 or 4 and a context of 131,072 tokens. The open-weight layer authors of \Cref{fig:author}, Ministral 3B, 8B and 14B \citep{liu2026ministral} and Mistral Small 3.2 24B\footnote{\url{https://huggingface.co/mistralai/Mistral-Small-3.2-24B-Instruct-2506}}, are served the same way on two A100 GPUs.

\paragraph{Artifacts.} We use the benchmarks as their creators intend, for evaluation only. DataClawBench and FDABench are released under the MIT license and HybridDeepResearch under Apache-2.0; the BIRD \citep{li2023bird} and Spider \citep{yu2018spider} databases that FDABench draws on are under CC BY-SA 4.0, and KramaBench's dataset card lists a Creative Commons license. The tasks are in English; DataClaw's data, which its creators de-identified, includes entity names in Chinese. We collect no data about people, and the hand checks of task-demand labels were done by the authors. The semantic layers, run traces and labels we derive are intended for research use.

\paragraph{Time per task.} Pooled over the tasks of all five benchmarks and all eight configurations, solving one task with an API model takes a median of 6.6 minutes and 12 model calls: 4.5 minutes and 7 calls for a single agent, and 9 minutes and 17 calls with orchestration. By benchmark, the median ranges from 2.1 minutes and 5 calls on KramaBench to 9.1 minutes and 15 calls on FDABench reports. Solving one task with GLM-4.7-Flash takes a median of 25 model calls, from 15 on FDABench multiple-choice to 47 on DataClaw. A semantic layer is written once per benchmark, in one pass of the author over every source: the DataClaw layer took about 1,800 calls and 17 minutes with GPT-5.5, and 31 to 58 minutes with the smaller authors; the KramaBench and FDABench layers, with about 47,000 and 350,000 entries, took several hours each. \Cref{app:cost} gives the cost per task in dollars.

\end{document}